\documentclass[accepted]{uai2021} %
\usepackage[T1]{fontenc}
\usepackage{inputenc}
\usepackage{verbatim}

\usepackage{tgtermes}
\usepackage{bm,amsbsy}
\usepackage{amsmath,amsfonts,amsthm,amssymb}
\usepackage[noend]{algpseudocode}
\usepackage{scalefnt,letltxmacro}
\LetLtxMacro{\oldtextsc}{\textsc}
\renewcommand{\textsc}[1]{\oldtextsc{\scalefont{1.10}#1}}
\usepackage[scaled=0.92]{PTSans}

\usepackage[usenames,dvipsnames]{xcolor}
\definecolor{shadecolor}{gray}{0.9}

\usepackage[colorinlistoftodos,
           prependcaption,
           textsize=small,
           backgroundcolor=yellow,
           linecolor=lightgray,
           bordercolor=lightgray
           ]{todonotes}

\usepackage{lineno}

\usepackage{ragged2e}

\DeclareRobustCommand{\parhead}[1]{\textbf{#1}~}

\usepackage{graphicx}
\usepackage[labelfont=bf]{caption}
\usepackage[format=hang]{subcaption}
\usepackage{pstricks, pst-node}
\usepackage{floatrow}

\usepackage{booktabs}
\usepackage{multirow}
\usepackage{float}

\usepackage{paralist}
\usepackage{enumitem} 

\usepackage[algoruled,linesnumbered]{algorithm2e}
\usepackage{listings}
\usepackage{fancyvrb}
\fvset{fontsize=\normalsize}

\usepackage[all]{hypcap}
\hypersetup{colorlinks}
\hypersetup{linktoc=all}
\hypersetup{citecolor=Violet}
\hypersetup{linkcolor=black}
\hypersetup{urlcolor=MidnightBlue}

\theoremstyle{plain}

\theoremstyle{plain}

\theoremstyle{plain}

\theoremstyle{plain}

\providecommand{\assumptionname}{Assumption}
\providecommand{\lemmaname}{Lemma}
\providecommand{\propositionname}{Proposition}
\providecommand{\theoremname}{Theorem}

\usepackage[nameinlink]{cleveref}
\creflabelformat{equation}{#1#2#3}
\Crefname{equation}{Eq.}{Eqs.}
\Crefname{assumption}{Assumption}{Assumptions}
\Crefname{prop}{Proposition}{Propositions}
\Crefname{lem}{Lemma}{Lemmas}

\usepackage[acronym,nowarn]{glossaries}
\usepackage{listings}
\lstdefinestyle{alp_style}{
    commentstyle=\color{OliveGreen},
    numberstyle=\tiny\color{black!60},
    stringstyle=\color{BrickRed},
    basicstyle=\ttfamily\scriptsize,
    breakatwhitespace=false,
    breaklines=true,
    captionpos=b,
    keepspaces=true,
    numbers=none,
    numbersep=5pt,
    showspaces=false,
    showstringspaces=false,
    showtabs=false,
    tabsize=2
}
\usepackage{xr}
\makeatletter
\newcommand*{\addFileDependency}[1]{
  \typeout{(#1)}
  \@addtofilelist{#1}
  \IfFileExists{#1}{}{\typeout{No file #1.}}
}
\makeatother

\DeclareRobustCommand{\trace}[1]{\textrm{Tr}}

\newacronym{AIS}{AIS}{annealed importance sampling}
\newacronym{DISIR}{DISIR}{dependent iterated sampling importance resampling}
\newacronym{ELBO}{ELBO}{evidence lower bound}
\newacronym{ESS}{ESS}{effective sample size}
\newacronym{HMC}{HMC}{Hamiltonian Monte Carlo}
\newacronym{ISIR}{ISIR}{iterated sampling importance resampling}
\newacronym{IWAE}{IWAE}{importance weighted auto-encoder}
\newacronym{KL}{KL}{Kullback-Leibler}
\newacronym{MCMC}{MCMC}{Markov chain Monte Carlo}
\newacronym{MH}{MH}{Metropolis-Hastings}
\newacronym{ML}{ML}{maximum likelihood}
\newacronym{MLP}{MLP}{feed forward neural network}
\newacronym{PIMH}{PIMH}{particle independent Metropolis-Hastings}
\newacronym{PPCA}{PPCA}{probabilistic principal component analysis}
\newacronym{SMC}{SMC}{sequential Monte Carlo}
\newacronym{VAE}{VAE}{variational auto-encoder}
\newacronym{VI}{vi}{variational inference}

\usepackage{amsmath,amsfonts,bm}

\def\eqref#1{equation~\ref{#1}}

\def\1{\bm{1}}

\DeclareMathAlphabet{\mathsfit}{\encodingdefault}{\sfdefault}{m}{sl}
\SetMathAlphabet{\mathsfit}{bold}{\encodingdefault}{\sfdefault}{bx}{n}

\usepackage[american]{babel}
\usepackage{natbib} %
\usepackage{mathtools} %
\usepackage{booktabs} %
\usepackage{tikz} %

\newcommand{\comm}[1]{} 
\newcommand{\bfm}{\mathbf{m}}

\newcommand{\bk}{\mathbf{k}}
\newcommand{\val}{v}
\newcommand{\feat}{{\phi}}
\newcommand{\Feat}{{\Phi}}
\newcommand{\taskparams}{{\psi}}
\newcommand{\tr}{t}
\newcommand{\data}{D}
\newcommand{\Exp}{\mathbb{E}}

\newcommand{\bff}{\mathbf{f}}

\newcommand{\bb}{\mathbf{b}}

\title{Information Theoretic Meta Learning with Gaussian Processes}

\author[1]{
Michalis~K.~Titsias} %
\author[1]{Francisco~J.~R.~Ruiz}
\author[2]{Sotirios Nikoloutsopoulos}
\author[1]{Alexandre Galashov}
\affil[1]{%
   DeepMind
}
\affil[2]{%
Athens Univ.\ of Econ.\ and Business, Greece
}

\begin{document}
\maketitle

\begin{abstract}
We formulate meta learning using information theoretic concepts; namely, mutual information and the information bottleneck. The idea is to learn a stochastic representation or encoding of the task description, given by a training set, that is highly informative about predicting the validation set. By making use of variational approximations to the mutual information, we derive a general and tractable framework for meta learning. This framework unifies existing gradient-based algorithms and also allows us to derive new algorithms. In particular, we develop a memory-based algorithm that uses Gaussian processes to obtain non-parametric encoding representations. We demonstrate our method on a few-shot regression problem and on four few-shot classification problems, obtaining competitive accuracy when compared to existing baselines.
\end{abstract}

\glsresetall

\section{Introduction}
\label{sec:introduction}

Meta learning \citep{Ravi2017OptimizationAA, Vinyals16,Edwards17,finn17a, lacoste2019uncertainty, nichol2018firstorder} and few-shot learning \citep{Fei-FeiFP06, Lake2011OneSL} aim at deriving data-efficient learning algorithms that can rapidly adapt to new tasks. To achieve that, these algorithms train deep neural networks on a set of tasks drawn from a common distribution. Each task is typically divided into a training (or support) set and a validation (or target) set. The neural network is fitted to predict each task's validation set from its support set. By sharing information across tasks, the network learns to rapidly adapt to new tasks and generalize from few examples at test time.

There is a plethora of work on few-shot learning algorithms, including memory-based \citep{Vinyals16, Ravi2017OptimizationAA} and gradient-based \citep{finn17a, nichol2018firstorder} procedures. Among those, the gradient-based model agnostic meta learning (MAML) by \citet{finn17a} has been particularly influential. However, despite the success of these algorithms, meta learning still lacks unifying principles that allow us to relate all these approaches and invent new schemes. While there exist probabilistic interpretations of existing methods, such as the approximate Bayesian inference approach \citep{grant2018recasting, Finn18, BMAML18} and the related conditional probability modelling 
approach \citep{garnelo18a,gordon2018metalearning}, there is not a general and tractable learning principle for meta learning that can help to get a better understanding of existing algorithms and derive new ones.

We address this issue in this paper. Specifically, we introduce an information theoretic view of meta learning, and we derive a general (and practical) framework for meta learning by exploiting variational approximations of the information theoretic quantities. We show that this framework recovers gradient-based algorithms, such as MAML and its probabilistic interpretations, and that it also allows us to derive new methods: we use it to introduce a new memory-based algorithm for supervised few-shot learning.

More in detail, we consider the information bottleneck principle \citep{tishby99information}, which can learn a stochastic encoding of the support set of each task that is highly informative for predicting the validation set of that task. The stochastic encoding is optimized through the difference between two mutual informations \citep{coverthomas2006}, so that the encoding compresses the training set into a representation that predicts well the validation set. Since the mutual information is intractable, we exploit recent approximations of the information bottleneck \citep{Alemietal2017,Chalk16, aless2016information} that make use of variational bounds \citep{barberagakov03}; this results in a tractable objective function for meta learning called variational information bottleneck (VIB).

\begin{figure*}[t]
    \centering
    \includegraphics[scale=0.48]{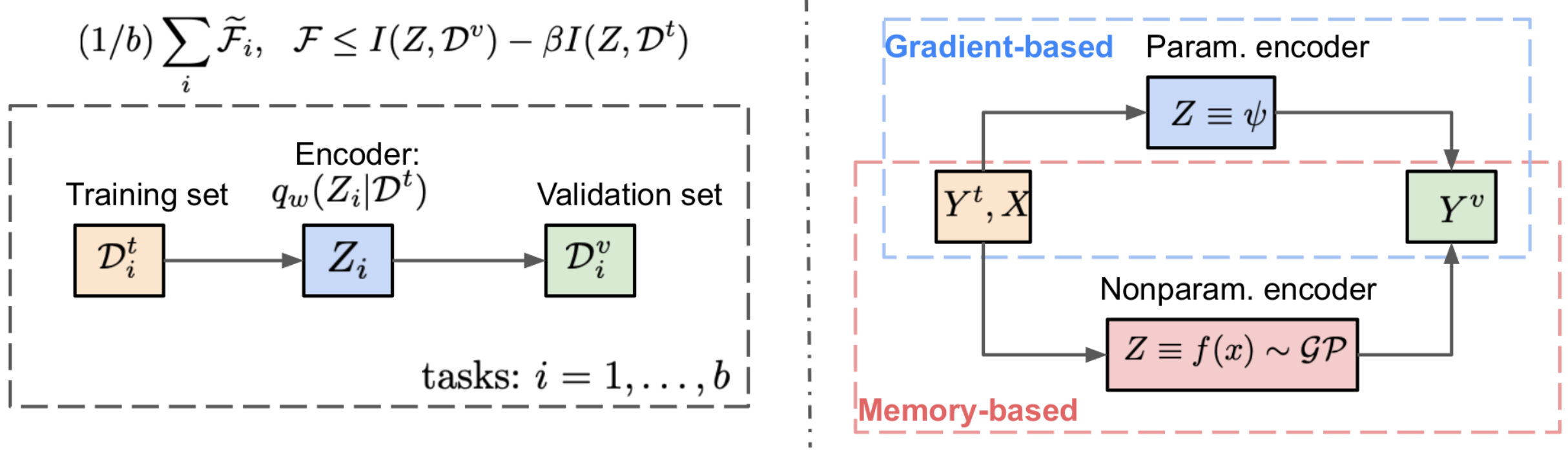}
    \vspace*{-10pt}
    \caption{({\bf Left}) Meta learning with the information bottleneck. The encoder $q_w(Z_i | \data^\tr_i)$ learns to compress each task's training set $\data^\tr_i$ by minimizing the mutual information $I(Z_i,\data^\tr_i)$ and to predict the task's validation set $\data^\val_i$ by maximizing $I(Z_i,\data^\val_i)$ (\Cref{sec:ib}). ({\bf Right}) Particularization to supervised few-shot learning, where we have input-output data for each task. MAML and GP memory-based methods (developed in this paper) are instances of this framework (\Cref{sec:stochasticMAML,sec:gpmetalearning}).}
    \label{fig:vib}
\end{figure*}

We show that VIB gives rise to gradient-based meta learning methods, such as MAML, when the encoding is \emph{parametric}, i.e., it has some model parameters or neural network weights. Additionally, VIB recovers a probabilistic version of MAML simply by using a stochastic encoding.
Furthermore, we use VIB to develop a memory-based algorithm for supervised few-shot learning (right panel in \Cref{fig:vib}), based on Gaussian processes (GPs)  \citep{rasmussen2006gaussian} and deep neural kernels \citep{wilson2016deep}, that offers a kernel-based Bayesian view of a memory system. With GPs, the underlying encoding takes the form of a \emph{non-parametric} function that follows a stochastic process amortized by the training set.
Our framework is general and would naturally allow for other extensions, such as combinations of memory and gradient-based meta learning.

We demonstrate our GP-based method on a few-shot regression problem and four classification problems. The former is a few-shot sinusoid regression problem, where we obtain smaller error than MAML. For the latter, we use the unified training and evaluation protocol of \citet{patacchiola2019deep}, obtaining that the GP-based algorithm provides competitive accuracy when compared to existing baselines, obtaining state-of-the-art results in some cases.

\section{Information Bottleneck View of Meta Learning}
\label{sec:ib}

We wish to learn from a distribution 
of tasks. During training, we observe a set of tasks, each consisting of a \emph{task description} represented by the support or training set $\data^\tr$ and a \emph{task validation} represented by the target or validation set $\data^\val$. At test time, we only have access to the training set $\data^\tr$ of a new task, and the goal of the algorithm is to adapt and form predictions on that task's $\data^\val$ or on further test data.

We formulate meta learning  using information theoretic concepts. The idea is to learn a stochastic representation or encoding of the task description $\data^\tr$ that is highly informative for predicting $\data^\val$. We introduce a random variable $Z$ that represents the encoding; it is drawn from a distribution $q_w(Z| \data^\tr)$  parameterized by $w$. Thus, the joint distribution over $\data^\tr$, $\data^\val$, and $Z$ is
\begin{equation}
    \label{eq:joint}
    q_w(\data^\val, \data^\tr, Z ) =
    q_w(Z| \data^\tr)
    p(\data^\val, \data^\tr),
\end{equation}
where $p(\data^\val, \data^\tr)$ denotes the unknown data distribution.

The goal of meta learning under this view is to tune the parameters $w$ of the encoder. To that end, one approach is to maximize the mutual information $I (Z, \data^\val)$ between $Z$ and the target set $\data^\val$. A trivial way to obtain a maximally informative representation is to set $Z=\data^\tr$, for which $I(Z, \data^\val)$ attains its maximum value $I(\data^\tr,\data^v)$. However, this is not a useful representation since 
 $Z$ simply memorizes the training set
and there is no learning involved. 
Instead, the information bottleneck (IB) principle \citep{tishby99information} adds a model complexity penalty to the maximization of $I (Z, \data^\val)$ 
to promote a more  compressive or parsimonious 
encoding of the training set $\data^\tr$, i.e., an encoding that extracts only relevant information from $\data^\tr$ for predicting $\data^\val$. 
The penalty term is the mutual information 
$I(Z, \data^\tr)$ that we want to 
minimize. %
Thus, the IB objective is
\begin{equation}
    \label{eq:L_IB}
    \mathcal{L}_{\textrm{IB}}(w) =  I(Z, \data^\val) - \beta I(Z, \data^\tr),
\end{equation}
where $\beta \geq 0$ is a hyperparameter. In this IB objective, 
the mutual information  $I(Z, \data^\tr)$ acts as a regularizer in the maximization of $I(Z, \data^\val)$, and it tries to introduce some partial independence between $Z$ and $\data^\tr$ so that irrelevant 
information is removed when encoding $\data^{\tr}$ into 
$Z$.

However, the IB objective is intractable because both mutual information terms depend on the unknown data distribution $p(\data^\val, \data^\tr)$. We overcome this by using variational bounds of the mutual information; in particular we obtain a tractable lower bound on $\mathcal{L}_{\textrm{IB}}(w)$ by lower bounding $I(Z, \data^\val)$ and upper bounding $I(Z, \data^\tr)$. The approach is similar to the one by \citet{Alemietal2017}, who introduced VIB for supervised learning of a single task.

\subsection{Variational Information Bottleneck (VIB) for Meta Learning} 
\label{sec:vib}

Here we construct a lower bound $\mathcal{F} \leq \mathcal{L}_{\textrm{IB}}(w)$. We first lower bound the mutual information $I(Z, \data^\val) = \text{KL}\left[q_w(Z, \data^\val) || q_w(Z) p(\data^\val) \right]$, given by
\begin{align}
    \label{eq:firstmutual}
    I(Z, \data^\val) & =
    \Exp_{q_w(Z, \data^\val)} 
    \left[\log 
    \frac{q_w(\data^\val| Z)}
    { p(\data^\val ) }\right],
\end{align}
where $\text{KL}$ denotes the Kullback-Leibler divergence and $q_w(\data^\val| Z) = \frac{\int q_w(Z| \data^\tr) p(\data^\val, \data^\tr) d \data^\tr}{ \int q_w(Z| \data^\tr)  p(\data^\val, \data^\tr) d \data^\tr d \data^\val}$ is intractable because it involves the data distribution $p(\data^\val, \data^\tr)$. To lower bound $I(Z, \data^\val)$, we follow \cite{barberagakov03} (see \Cref{app:boundsonMI}) and introduce a \emph{decoder} model $p_\theta(\data^\val | Z)$ to approximate the intractable $q_w(\data^\val| Z)$, where $\theta$ are additional parameters, yielding the bound\footnote{%
    The bounds are valid even when the parameters $w$ of the encoder $q_w(Z| \data^\tr)$ and $\theta$ of the decoder $p_\theta(\data^\val| Z)$ and prior $p_\theta(Z)$ have shared components (e.g., shared neural network weights).
}
\begin{equation}
    \label{eq:firstmutual_bound}
    \begin{split}
        I(Z, \data^\val) 
        & \geq \Exp_{q_w(Z, \data^\val)} 
        \left[\log 
        \frac{p_\theta(\data^\val| Z)}
        {p(\data^\val ) } \right] \\  
        & = \Exp_{q_w(Z, \data^\val)} \left[
        \log  p_\theta(\data^\val| Z) \right]  
        + \mathcal{H}(\data^\val),
    \end{split}
\end{equation}
where the entropy $\mathcal{H}(\data^\val)$ is a constant that does not depend on the tunable parameters $(\theta, w)$.
Secondly, we upper bound the mutual information $I(Z, \data^\tr)$,
\begin{align}
    I(Z, \data^\tr)
    = \Exp_{q_w(Z, \data^\tr)} 
    \left[ \log 
    \frac{q_w(Z | \data^\tr)}
    {q_w(Z) } \right],
\end{align}
where $q_w(Z) = \int q_w(Z | \data^\tr)p(\data^\tr) d \data^\tr$ is intractable because it involves the unknown data distribution $p(\data^\tr)$. Similarly, we approximate $q_w(Z)$ with a tractable \emph{prior} model distribution $p_\theta(Z)$, leading to the bound
\begin{align}
    \label{eq:secondmutual_bound}
    I(Z, \data^\tr) 
    & \leq \Exp_{q_w(Z, \data^\tr)} 
    \left[ \log 
    \frac{q_w(Z | \data^\tr)}
    {p_\theta(Z) } \right].
\end{align}
We obtain the overall bound, $\mathcal{F}(\theta, w) + \mathcal{H}(\data^\val) \leq \mathcal{L}_{\textrm{IB}}(w)$, by applying \Cref{eq:firstmutual_bound,eq:secondmutual_bound} into \Cref{eq:L_IB},
\begin{equation}
    \begin{split}
        \mathcal{F}(\theta,w)  
        = \;& \Exp_{q_w(Z, \data^\val) } \left[\log  p_\theta(\data^\val| Z) \right] \\
        & - \beta \Exp_{q_w(Z, \data^\tr)}   
        \left[\log \frac{q_w(Z | \data^\tr)}
        {p_\theta(Z) } \right],
    \end{split}
\end{equation}
where we have dropped the constant $\mathcal{H}(\data^\val)$ from the objective function $\mathcal{F}(\theta,w)$.

Thus, given a set of $b$ tasks $\{\data^\tr_i, \data^\val_i \}_{i=1}^b$, learning the parameters $(\theta, w)$ during meta training reduces to maximizing the empirical average $\frac{1}{b} \sum_{i} \mathcal{\widetilde{F}}_i(\theta,w)$, where each $\mathcal{\widetilde{F}}_i(\theta,w)$ is an unbiased estimate of $\mathcal{F}(\theta,w)$ based on the $i$-th task, given by (see \Cref{sec:generalcase})
\begin{equation}
    \label{eq:vibobjective}
    \begin{split}
        \mathcal{\widetilde{F}}_i(w,\theta) 
        = & \; \Exp_{q_w(Z_i | \data^\tr_i)} \left[ \log p_\theta(\data^\val_i| Z_i) \right] \\  
        & -  \beta \text{KL}[q_w(Z_i | \data^\tr_i) || p_\theta(Z_i)].
    \end{split}
\end{equation}
The meta-training procedure is carried out in different episodes, where at each step we receive a minibatch of tasks and perform a stochastic gradient maximization step based on the data in that minibatch.

Note that the objective in \Cref{eq:vibobjective} is similar to the variational inference objectives for meta learning \citep{RaviB19}. In particular, it can be viewed as an evidence lower bound (ELBO) on the log marginal likelihood of the validation set,  $\log \int p_\theta(\data^\val_i| Z_i) p_\theta(Z_i) d Z_i$, with the following differences: (i) there is a hyperparameter $\beta$ in front of the KL term, and (ii) the distribution $q_w(Z_i | \data^\tr_i)$ is more restricted than in variational inference, since it now acts as a \emph{stochastic bottleneck} that encodes the support set $\data^\tr_i$ (i.e., it is amortized by $\data^\tr_i$) and via the term $\Exp_{q_w(Z_i | \data^\tr_i)} [\log p_\theta(\data^\val_i| Z_i)]$ it is optimized to reconstruct the validation set.

\subsection{Information Theoretic View of MAML-type Methods} 
\label{sec:stochasticMAML}

We recover MAML \citep{finn17a} as a special case of the VIB framework. To see this, suppose that the encoding variable $Z_i$ for the $i$-th task coincides with a vector of some task-specific model parameters or neural network weights $\taskparams_i$, so that $p_\theta(\data^\val_i| Z_i) \equiv  p(\data^\val_i | \taskparams_i)$, and   $p_\theta(Z_i)\equiv p_\theta(\taskparams_i)$ is the prior over these parameters. MAML tries to find a shared initial parameter value $\theta$ so that few gradient steps based on the support set objective, $\log p(\data^\tr_i|\theta)$, lead to a task-specific parameter value $\taskparams_i$ with good predictive capacity on the validation set. That is, MAML estimates the task parameters by $\taskparams_i = \theta + \Delta(\theta,\data^\tr_i)$, where $\Delta(\theta,\data^\tr_i)$ denotes these inner loop adaptation steps---for one step of stochastic gradient descent (SGD) with step size $\rho$, it is just $\rho \nabla_{\theta} \log p(\data^\tr_i|\theta)$. We recover MAML from \Cref{eq:vibobjective} by setting $\beta=0$ and setting the encoder to a deterministic Dirac delta measure, $q_\theta(\taskparams_i | \data^\tr_i)=\delta_{\theta + \Delta(\theta,\data^\tr_i)}(\taskparams_i)$, so that $\mathcal{\widetilde{F}}_i(\theta) = \log p(\data^\val_i |\theta + \Delta(\theta,\data^\tr_i))$.

\parhead{Bayesian or probabilistic MAML.}
Based on this view, we can generalize MAML by using a probabilistic encoder instead of a Dirac delta. For instance, we can use a Gaussian encoder, $q_{\theta,s}(\taskparams_i | \data^\tr_i) = \mathcal{N}(\taskparams_i | \theta + \Delta(\theta,\data^\tr_i), s)$, where $s$ is a diagonal covariance. Then, the objective 
becomes
\begin{align}
    \label{eq:stochmaml}
    \mathcal{\widetilde{F}}_i(\theta, s) 
    = & \; \Exp_{\mathcal{N}(\epsilon|0,I)} \left[ \log  p(\data^\val| \theta + \Delta(\theta,\data^\tr_i) + \sqrt{s} \circ  \epsilon) \right] \nonumber \\
    & - \beta \text{KL}\left[ q_{\theta,s}(\taskparams_i | \data^\tr_i) || 
    p_\theta(\taskparams_i) \right],
\end{align}
where we have reparameterized the expectation following \citet{kingma2013auto} for stochastic optimization of the meta parameters $(\theta,s)$. 
This connects with several approaches in the literature  
 that have introduced probabilistic or Bayesian MAML 
 algorithms \citep{grant2018recasting, Finn18,BMAML18,nguyen2019uncertainty}. From the VIB perspective, such probabilistic MAML methods are associated with a probabilistic encoder that introduces uncertainty, where the hyperpameter $\beta$ in the VIB objective controls the amount of uncertainty.

\subsection{VIB for Supervised Meta Learning}
\label{sec:vibSuper}

Here, we explain how to adapt the VIB principle to supervised meta learning. Consider a few-shot supervised learning problem, where for each task we wish to predict outputs or labels given the corresponding inputs. In this supervised setting, we denote the task support set as $\data^\tr = (Y^\tr, X^\tr)$, where $Y^\tr = \{y^\tr_j\}_{j=1}^{n^\tr}$ and $X^\tr = \{x^\tr_j\}_{j=1}^{n^\tr}$ denote the output and input observations, respectively. Similarly, we write $\data^\val =  (Y^\val, X^\val)$ for the validation set. During meta testing, for any new task we observe the support set $\data^\tr_* =(Y_*^\tr, X^\tr_*)$ together with the test inputs $X^v_*$ and the goal is to predict the test outputs $Y^\val_*$.

This suggests that we can construct a task encoder distribution of the form $q_w(Z| Y^\tr, X)$ that depends on the training outputs $Y^\tr$ and generally on all inputs $X = (X^\tr, X^\val)$.\footnote{%
    Dependence on all inputs allows explaining both transductive and non-transductive settings \citep{bronskill2020tasknorm, finn17a, nichol2018firstorder} as special cases (see \Cref{app:transductive}).
} 
We would like to train this encoder so that $Z$ becomes highly predictive about the validation outputs $Y^\val$ and simultaneously compressive about $Y^\tr$. Then, we form a VIB objective based on the \emph{input-conditioned} information bottleneck, $I(Z,Y^\val|X) - \beta I(Z,Y^\tr|X)$, i.e., where both mutual information terms are conditional. Similarly to \Cref{sec:vib}, we lower bound this objective and approximate it by an unbiased empirical average, 
$\frac{1}{b} \sum_{i=1}^b 
\mathcal{\widetilde{F}}_i(\theta,w)$, where
\begin{equation}
    \label{eq:VIBsuper}
    \begin{split}
        \mathcal{\widetilde{F}}_i(\theta,w)  
        = & \; \Exp_{q_w(Z_i| Y^\tr_i, X_i)} 
        \left[\log p_\theta(Y^\val_i | Z_i, X_i ) \right] \\
        & -  \beta 
         \text{KL}\left[ q_w(Z_i | Y^\tr_i, X_i) 
        || 
        p_\theta(Z_i|X_i) \right].
    \end{split}
\end{equation}
Here, $p_\theta(Y^\val_i| X^\val_i, Z_i)$ and $p_\theta(Z_i|X_i)$ are the decoder and prior model distributions introduced by the variational approximation (see \Cref{app:vibsuper} for a detailed derivation).

\Cref{eq:VIBsuper} provides the VIB objective for supervised meta learning. We can recover the supervised version of MAML as a special case of VIB, similarly to \Cref{sec:stochasticMAML}.
\section{Supervised Meta Learning with Gaussian Processes}
\label{sec:gpmetalearning}

In this section, we use the VIB framework for meta learning to develop a new algorithm for few-shot supervised learning, which uses a non-parametric stochastic encoder based on a GP model. In \Cref{sec:gpmemory} we outline the structure of the GP meta learning method and in \Cref{sec:amortisationq} we give further details about how to set up the GP encoder.

\subsection{Gaussian Process VIB-based method}
\label{sec:gpmemory}

As described in \Cref{sec:vibSuper}, the VIB framework for few-shot supervised learning requires us to specify the encoding variable $Z_i$, together with the encoder $q_w(Z_i| Y^\tr_i, X_i)$, the decoder over the validation outputs $p_\theta(Y^\val_i| Z_i, X_i)$, and the prior model $p_\theta(Z_i|X_i)$. Here, we construct these quantities using a GP model \citep{rasmussen2006gaussian} in order to obtain flexible non-parametric stochastic functions.

\parhead{GP specification.}
We introduce an (unknown) task-specific function $f_i(x)$ that is \emph{a priori} (before observing any task data) drawn from a GP, i.e., $f_i(x) \sim GP(0, k_{\theta}(x, x'))$, where $k_\theta$ denotes the kernel function. Without loss of generality, we use a deep kernel function, $k_{\theta}(x, x') = \sigma^2_f  \feat(x;\theta)^\top \feat(x';\theta)$, where $\feat(x;\theta)$ is a feature vector given by a deep neural network parameterized by $\theta$, and $\sigma^2_f$ is the kernel variance parameter (which, if learnable, we also consider to be part of the full set of parameters $\theta$).
Equivalently, we can interpret this construction as setting $f_i(x) = \feat(x;\theta)^\top \theta^{\textrm{out}}_i$, i.e., a linear function of the feature vector with task-specific weights drawn from a Gaussian distribution, $\theta^{\textrm{out}}_i \sim \mathcal{N}(\theta^{\textrm{out}}_i |0, \sigma_f^2 I )$. In the GP formulation, the weights $\theta^{\textrm{out}}_i$ are marginalized out and we are left with the parameters $\theta$ shared across tasks.

Suppose now that we observe the task data, i.e., the support $\data^\tr_i = (Y^\tr_i, X^\tr_i)$ and validation $\data^\val_i = (Y^\val_i, X^\val_i)$ sets, so that we can evaluate the task function $f_i(\cdot)$ on all task inputs $X_i = (X^\tr_i, X^\val_i)$. Let $f^\val_{i,j} \equiv f(x^\val_{i,j})$ denote the function value at the validation input $x^\val_{i,j}$ associated with  output $y^\val_{i,j}$, where the index $j$ runs over the instances in $\data^\val_i$. Let $\bff^\val_i = \{f^\val_{i,j}\}_{j=1}^{n^\val}$ be the vector containing all such values. Similarly, let $\bff^\tr_i$ be the vector of function values at the inputs $X^\tr_i$.

\parhead{Components of the VIB formulation.}
We now specify the four ingredients of VIB: (i) the encoding variable $Z_i$, (ii) the prior $p_\theta(Z_i|X_i)$, (iii) the decoder $p_\theta(Y^\val_i| Z_i, X_i)$, and (iv) the encoder $q_w(Z_i| Y^\tr_i, X_i)$.
\textbf{(i)} We set the task encoding variable $Z_i$ to the full set of function values, $Z_i \equiv (\bff^\val_i, \bff^\tr_i)$. It is a \emph{non-parametric} encoding, since its size grows with the number of task data points.
\textbf{(ii)} We set the prior model to the GP prior, $p_\theta(Z_i|X_i) \equiv p(\bff^\val_i, \bff^\tr_i|X_i)$, where\footnote{%
    We drop the dependency on $\theta$ for notational simplicity.
}
\begin{align}
    \label{sec:GPpriorjoint}
    & p(\bff^\val_i, \bff^\tr_i|X_i)
    = p(\bff^\val_i | \bff^\tr_i, X_i) \times
    p(\bff^\tr_i|X^\tr_i), \\
    & = 
    \mathcal{N}(\bff^\val_i| {\bf K}_i^{\val \tr} 
    [{\bf K}^{\tr}_i]^{-1} \bff^\tr_i,  {\bf K}^\val_i - {\bf K}_i^{\val \tr} [{\bf K}^{\tr}_i]^{-1}  [{\bf K}^{\val \tr}_i]^\top)  \nonumber \\
    & \quad \times  \mathcal{N} (\bff^\tr_i | {\bf 0}, {\bf K}^\tr_i). \nonumber
\end{align}
Here, ${\bf K}^{\tr}_i$ and ${\bf K}^{\val}_i$ are $n^\tr \times n^\tr$ and $n^\val \times n^\val$ kernel matrices on the training $X^\tr_i$ and validation inputs $X^\val_i$, respectively, and ${\bf K}^{\val \tr}_i$ is the $n^\val \times n^\tr$ cross kernel matrix between the two sets of inputs.
\textbf{(iii)} We set the decoder model $p_\theta(Y^\val_i | \bff^\val_i, \bff^\tr_i, X_i)$ to the standard GP likelihood. For i.i.d.\ observations, $Y^\val_i$ is independent of $\bff^\tr_i$ and $X_i$ given $\bff^\val_i$, and the likelihood factorizes across data points, $p(Y^\val_i | \bff^\val_i) = \prod_{j=1}^{n^\val} p(y^\val_{i,j} | f^\val_{i,j})$. Each $p(y^\val_{i,j} | f^\val_{i,j})$ is a standard likelihood model, such as a Gaussian density $p(y^\val_{i,j} | f^\val_{i,j}) = \mathcal{N}(y^\val_{i,j} | f^\val_{i,j}, \sigma^2)$ for regression problems or a categorical/softmax likelihood for few-shot classification (see \Cref{app:multiclass}).
\textbf{(iv)} We set the encoder $q_w(Z_i| Y^\tr_i, X_i) \equiv q(\bff^\val_i, \bff^\tr_i| Y^\tr_i, X_i)$ as the product
\begin{equation}
    \label{eq:gpencoder} 
    q(\bff^\val_i, \bff^\tr_i|Y^\tr_i, X_i)
    =  p(\bff^\val_i| \bff^\tr_i, X_i) 
    q(\bff^\tr_i|\data^\tr_i),
\end{equation}
where $p(\bff^\val_i| \bff^\tr_i|X_i)$ is the conditional GP prior from \Cref{sec:GPpriorjoint}, and $q(\bff^\tr_i|\data^\tr_i)$ is an encoder of the training set that takes the form of a Gaussian distribution amortized by $\data^\tr_i$; see \Cref{sec:amortisationq} for details. \Cref{eq:gpencoder} shares a similar structure with a standard posterior GP, where we first observe the training set $\data^\tr_i$, then we compute the (approximate) posterior $q(\bff^\tr_i|\data^\tr_i)$, and finally we predict the validation set function values at inputs $X^\val_i$ based on the conditional GP prior $p(\bff^\val_i| \bff^\tr_i, X_i)$.

\parhead{Objective function.}
Putting all together, we obtain the following VIB single-task objective (see \Cref{app:dervibbound}),
\begin{equation}
    \label{eq:vibgpobjective}
    \sum_{j=1}^{n^\val} \! \Exp_{q(f^\val_{i,j})}\! 
    [\log  p(y^\val_{i,j} | f^\val_{i,j}) ] \!-\!  \beta 
    \text{KL}\! \left[ q(\bff^\tr_i | \data^\tr_i) || p(\bff^\tr_i | X^\tr_i) \right]\!\!,
\end{equation}
where $q(f^\val_{i,j}) = \int p(f^\val_{i,j} |\bff^\tr_i, x^\val_{i,j}, X^\tr_i) q(\bff^\tr_i | \data^\tr_i) d \bff^\tr_i$ is a univariate Gaussian distribution over one validation function value $f^\val_{i,j}$. The distribution $q(f^\val_{i,j})$ depends on the training set and the validation input $x^\val_{i,j}$. Thus, to compute it, we simply use the univariate conditional GP $p(f^\val_{i,j} |\bff^\tr_i, x^\val_{i,j}, X^\tr_i)$ to extrapolate the function value at $x^\val_{i,j}$.

\subsection{GP Encoder %
}
\label{sec:amortisationq}

We now specify the encoder $q(\bff^\tr| \data^\tr)$ used in \Cref{eq:gpencoder}. A suitable choice is to set it equal to the exact posterior distribution over $\bff^\tr$ given the training set, i.e., $q(\bff^\tr| \data^\tr)=p(\bff^\tr | \data^\tr) \propto \prod_{j=1}^{n^\tr} p(y^\tr_j | f^\tr_j) \mathcal{N}(\bff^\tr | 0,  {\bf K}^\tr)$.

The posterior $p(\bff^\tr | \data^\tr)$ is tractable for standard regression problems with Gaussian likelihood, i.e., $p(y^\tr_j| f^\tr_j) = \mathcal{N}(y^\tr_j | f^\tr_j, \sigma^2)$. In this case, the posterior is also Gaussian,
\begin{align}
    \label{eq:pft}
    & p(\bff^\tr  |  \data^\tr) \!
    \\
    & = \! \mathcal{N}\!(\bff^\tr  | 
    {\bf K}^\tr(  {\bf K}^\tr + \sigma^2 I)^{-1} Y^\tr, 
     {\bf K}^\tr \! - \! {\bf K}^\tr({\bf K}^\tr + \sigma^2 I)^{-1}  {\bf K}^\tr), \nonumber
\end{align}
and thus the encoder $q(\bff^\tr | \data^\tr) = p(\bff^\tr | \data^\tr)$ is tractable. Note that $q(\bff^\tr | \data^\tr)$ depends on $\data^\tr$ through the kernel matrix ${\bf K}^\tr$, which mixes all the training inputs in $X^\tr$. The posterior covariance depends only on $X^\tr$, while its mean depends additionally on $Y^\tr$. This choice of the encoder does not require to introduce any extra variational parameters $w$, as $q(\bff^\tr | \data^\tr)$ depends only on the model parameters $\theta$ that appear in the kernel function and (possibly) in the likelihood.

When the likelihood is not Gaussian, the posterior $p(\bff^\tr | \data^\tr)$ is not available analytically. In this case, we set $q(\bff^\tr| \data^\tr)$ to an approximate posterior. Specifically, we approximate each non-Gaussian likelihood term $p(y^\tr_j| f^\tr_j)$ with a Gaussian term; this is similar to the Gaussian approximations of variational Bayes or expectation-propagation in GPs \citep{rasmussen2006gaussian,opperarchambeau09,HensmanZL14}. That is, we approximate
\begin{equation}
    p(y^\tr_j | f^\tr_j )
    \approx \mathcal{N}(m^\tr_j | 
    f^\tr_j,
    s^\tr_j),
\end{equation}
where $m^\tr_j \equiv m_w(y^\tr_j, x^\tr_j) \in \mathbb{R}$ and $s^\tr_j \equiv s_w(y^\tr_j, x^\tr_j)  \in \mathbb{R}_+$ are amortized functions parameterized by $w$ that take as input a data point $(y^\tr_j, x^\tr_j)$ associated with the latent variable $f^\tr_j$ and output the parameters of the Gaussian approximation. In this case, the amortized encoder becomes a fully dependent multivariate Gaussian distribution of the form 
\begin{align}
    & q(\bff^\tr \! | \data^\tr) \nonumber \\
    & = \mathcal{N}(\bff^\tr
    | {\bf K}^\tr(  {\bf K}^\tr + {\bf S}^\tr )^{-1} \bfm^\tr, \! 
    {\bf K}^\tr \! - \! {\bf K}^\tr({\bf K}^\tr + {\bf S}^\tr)^{-1}  {\bf K}^\tr ), \nonumber 
\end{align}
where ${\bf S}^\tr$ is a diagonal covariance matrix with the vector $(s^\tr_1, \ldots, s^\tr_{n^\tr})$ in its diagonal, and $\bfm^\tr$ is the vector of values $(m^\tr_1, \ldots, m^\tr_{n^\tr})$. Using this encoder, we can re-write the VIB objective from \Cref{eq:vibgpobjective} in a form that is computationally more convenient (see \Cref{app:dervibbound}),  
\begin{align}
    & \sum_{j=1}^{n^\val} \Exp_{q(f^\val_j)}[ 
    \log  p(y^\val_j | f^\val_j)]  
     -  \beta \sum_{j=1}^{n^\tr} \Exp_{q(f^\tr_j)} 
    [\log \mathcal{N}(m^\tr_j | 
    f^\tr_j,s^\tr_j)] \nonumber \\
    & + \beta \log \mathcal{N}(\bfm^\tr |0, {\bf K}^\tr + {\bf S}^\tr), 
\end{align}
where each marginal Gaussian distribution $q(f_j)$ is computed using the same expression,
$
q(f_j)
= \mathcal{N}(f_j| \bk^\tr_j 
({\bf K}^\tr + {\bf S}^\tr)^{-1} \bfm^\tr
,k_j 
-  \bk^\tr_j ({\bf K}^\tr + {\bf S}^\tr)^{-1} {\bk^\tr_j}^\top)
$,
regardless of whether $x_j$ is from the validation or the training set (or any other further test set). Here, $\bk^\tr_j \equiv k(x_j,X^\tr)$ is the $n^\tr$-dimensional row vector of kernel values between $x_j$ and the training inputs $X^\tr$, and $k_j \equiv k(x_j,x_j)$. 

We can simplify the encoder by assuming a constant value $s_w(x_j^\tr) : = \sigma^2$ and further assuming that $m_w(y_j^\tr, x^\tr_j)$ depends only on the output $y_j^\tr$, i.e., $m_w(y_j^\tr, x^\tr_j) : = m_w(y^\tr_j)$. In our experiments, we found that this simplification worked better in most few-shot settings.

\parhead{GP encoder for classification problems.}
We now particularize the encoder for classification problems, which is the standard application in few-shot learning. For notational simplicity, we focus on binary classification, and we describe 
multi-class classification in \Cref{app:multiclass}.

Consider a meta learning problem in which each task is a binary classification problem. The class labels take value in $ \{-1,1\}$ and the (non-Gaussian) likelihood is the sigmoid $p(y^\tr_j | f^\tr_j) = 1 /(1 + e^{- y^\tr_j f^\tr_j })$.  
To specify the GP encoder, we simply
need to choose the form of
the amortized functions $(m_w(y^\tr_j, x^\tr_j), s_w(y^\tr_j, x_j^\tr))$. We set them as   
$$
m(y^\tr_j, x^\tr_j) : = y^\tr_j \times \widetilde{m}_w
(x^\tr_j),  \  \ s_w(y^\tr_j, x_j^\tr) : =  s_w(x_j^\tr), 
$$
where $\widetilde{m}_w(x^\tr_j)$ and $s_w(x_j^\tr)$ are parametrized by a neural network. Note that the dependence on the output label $y^\tr_j \in \{-1,1\}$ simply changes the sign of $\widetilde{m}_w(x^\tr_j)$,
which yields amortization invariance to class re-labeling. That is, if we swap the roles of the two labels $\{-1,1\}$, the amortization remains valid and it does not require any change.

We now describe the simplified encoder. We obtain it by assuming a constant form for the functions $(\widetilde{m}_w(x^\tr_j), s_w(x_j^\tr)) : = (\widetilde{m}, \sigma^2)$; this has significant computational benefits because it only requires to tune two scalar parameters $(\widetilde{m},\sigma^2)$. As mentioned above, the simplified encoder led to improved performance in our experiments.

\section{Related Work}
\label{sec:related}

In this work, we formulated meta learning using the VIB principle. VIB has been used before for different purposes, such as regularization of single-task supervised learning \citep{Alemietal2017}, sparse coding \citep{Chalk16}, re-interpretation of variational autoencoders \citep{burgess2018understanding} and dropout \citep{aless2016information}, and for compression of deep neural networks \citep{Dai18}.

A meta learning method that connects with the information bottleneck was recently proposed by \citet{hu2020empirical}. They use the information bottleneck to analyze the generalization of a variational Bayesian inference objective suitable for transductive supervised few-shot learning. The information bottleneck from Theorem 1 of  \citet{hu2020empirical} differs from the information bottleneck objective in our paper (the objective in \Cref{eq:VIBsuper} for the supervised learning case) in the second term.  This second term is the conditional entropy $\mathcal{H}(Z|D^\tr) = -E_{q(Z,D^\tr)}[\log q(Z|\data^\tr)]$  in Eq.~19 of \citet{hu2020empirical}, while it is a mutual information in our case, $I(Z,\data^\tr) = \mathcal{H}(Z) - \mathcal{H}(Z|\data^\tr)$. When used as a regularizer, the conditional entropy $\mathcal{H}(Z|\data^\tr)$ alone
may lead to simply increasing the variance of $q(Z|\data^\tr)$ and not extracting the relevant information from $\data^\tr$ to predict $\data^\val$. Moreover, this term is a differential entropy and thus it can be unbounded.
In contrast, $I(Z,\data^\tr)\geq 0$ is bounded, so it does not allow the variance of $q(Z|\data^\tr)$ to blow and it encourages extracting relevant information from $\data^\tr$ to predict $\data^\val$. 

Given the probabilistic nature of our framework, we can relate it to other probabilistic or Bayesian 
approaches, and particularly with those that: (i)  probabilistically re-interpret and extend gradient-based methods \citep{grant2018recasting, Finn18,BMAML18,nguyen2019uncertainty,gordon2018metalearning,chen2019modular} and (ii) derive amortized conditional probability models \citep{garnelo18a,gordon2018metalearning}. The underlying learning principle in both (i)-(ii) is to construct and maximize a predictive distribution (or conditional marginal likelihood) of the validation points given the training points, which, e.g., in supervised few-shot learning is written as $p_\theta(Y^\val | X^\val, X^\tr, Y^\tr) = \int p(Y^\val | X^\val, \taskparams_i) p_{\theta}(\taskparams_i | X^\tr, Y^\tr) d \taskparams_i = \frac{p_{\theta}(Y^\val, Y^\tr | X^\val, X^\tr) }{p_{\theta}(Y^\tr | X^\tr) }$. Here, $p_\theta(\taskparams_i | X^\tr, Y^\tr)$ is a posterior distribution over the task parameters $\taskparams_i$, after observing the training points, and $\theta$ is a meta parameter which for simplicity we assume to be found by point estimation. However, this objective is difficult to approximate. Unlike the marginal likelihood on all task outputs $p_\theta(Y^\val, Y^\tr | X^\val, X^\tr)$, for which we can easily compute a lower bound, there is no tractable lower bound on the predictive conditional $p_\theta(Y^\val | X^\val, X^\tr, Y^\tr)$.\footnote{%
    To obtain such a bound, we either need to have access to the intractable posterior $p_\theta(\taskparams_i | X^\tr, Y^\tr)$ or to upper bound the marginal likelihood on the training points $p_\theta(Y^\tr | X^\tr)$, which is hard.
} 
This inherent difficulty with computing the predictive distribution has led to several approximations, i.e., methods of category (i) above, ranging from maximum a posteriori (MAP), Laplace, variational inference procedures (without bounds on the predictive conditional) and Stein variational gradient descent. The approaches of category (ii) try to directly model $p_\theta(Y^\val | X^\val, X^\tr, Y^\tr)$ without considering this as an approximation to an initial joint Bayesian model. Our VIB framework differs significantly from the predictive distribution principle, since VIB has an information theoretic motivation and it rigorously bounds an information bottleneck objective. VIB is also a fully tractable objective, thus avoiding the need to choose a particular approximate inference method and allowing us to rather focus on setting up the encoding procedure, as demonstrated with the GP example from \Cref{sec:gpmetalearning}.

Finally, regarding related works of GPs in meta learning, the ALPaCA method \citep{harrison2018metalearning} applied GPs to Bayesian linear regression, while  \citet{tossou2019adaptive} used kernel-based methods from a regularization rather than Bayesian perspective. 
Closer to our work, \citet{patacchiola2019deep} and \citet{snell2020bayesian} used GPs with deep neural kernels for few-shot classification. Our usage of GPs is different; e.g., our encoder amortization strategy can potentially deal with arbitrary likelihood functions and task output observations, while \citet{patacchiola2019deep} assume a Gaussian likelihood for the binary class labels and  \citet{snell2020bayesian} consider the P\'{o}lya-gamma augmentation, which is tailored to classification problems.

\section{Experiments}
\label{sec:exps}

Here we evaluate the algorithm from  \Cref{sec:gpmetalearning} (labeled ``GP-VIB'' in this section). To that end, we consider a standard set of meta-learning benchmarks: sinusoid regression and few-shot classification.

\begin{figure*}[t]
    \centering
    \begin{tabular}{ccc}
    $K=1$ & $K=2$ & $K=3$ \\
    {\includegraphics[scale=0.22]
    {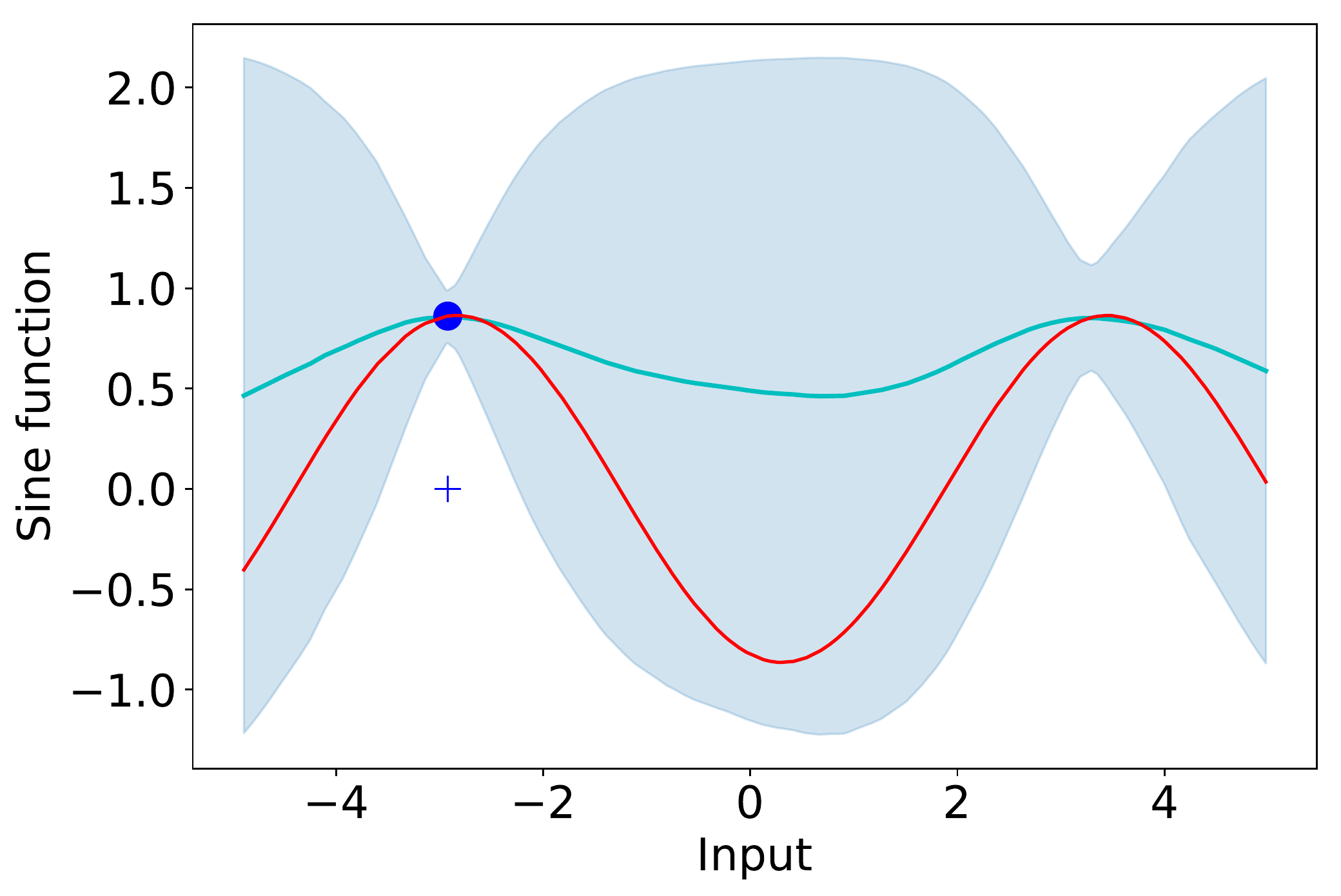}} &
    {\includegraphics[scale=0.22]
    {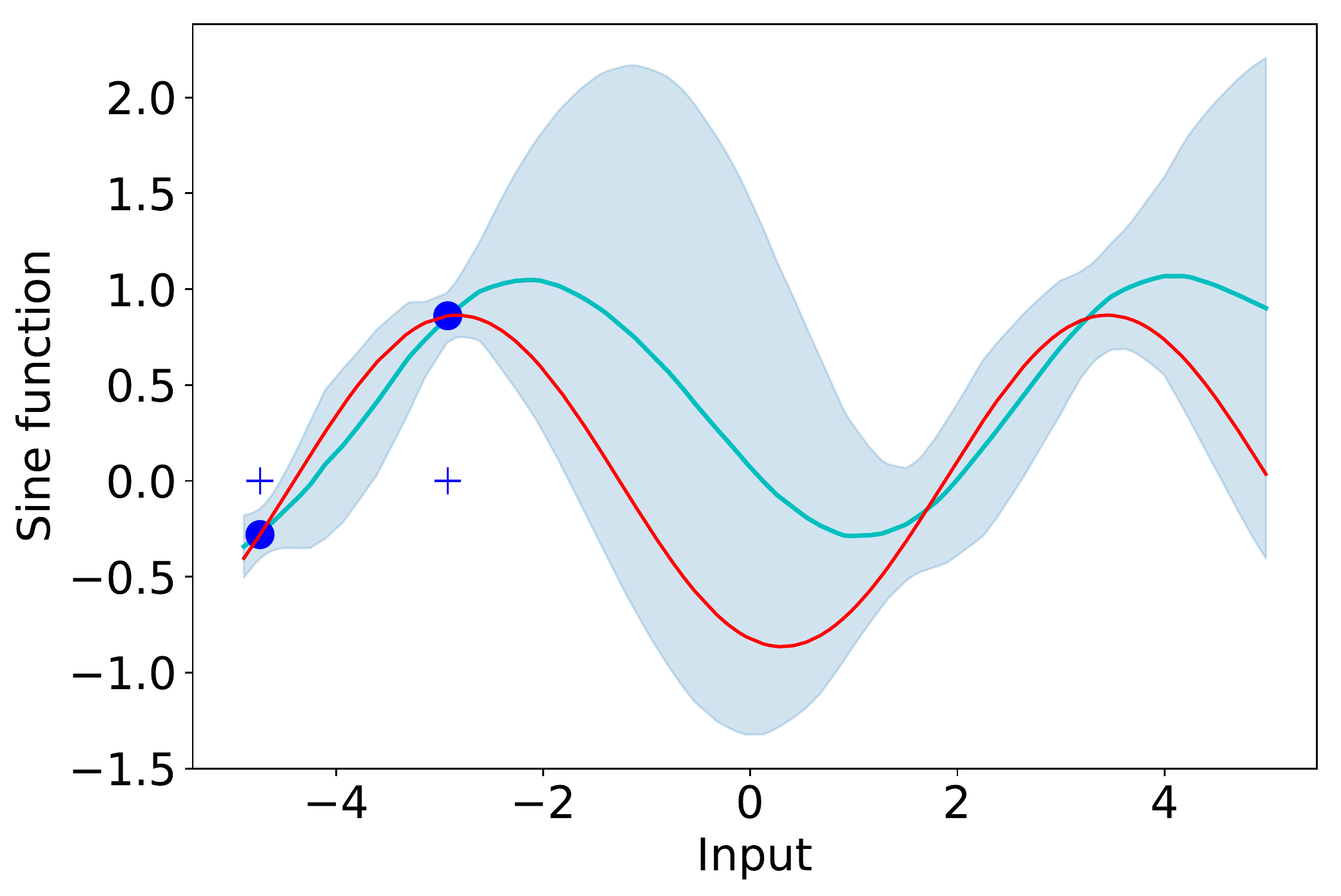}} &
    {\includegraphics[scale=0.22]
    {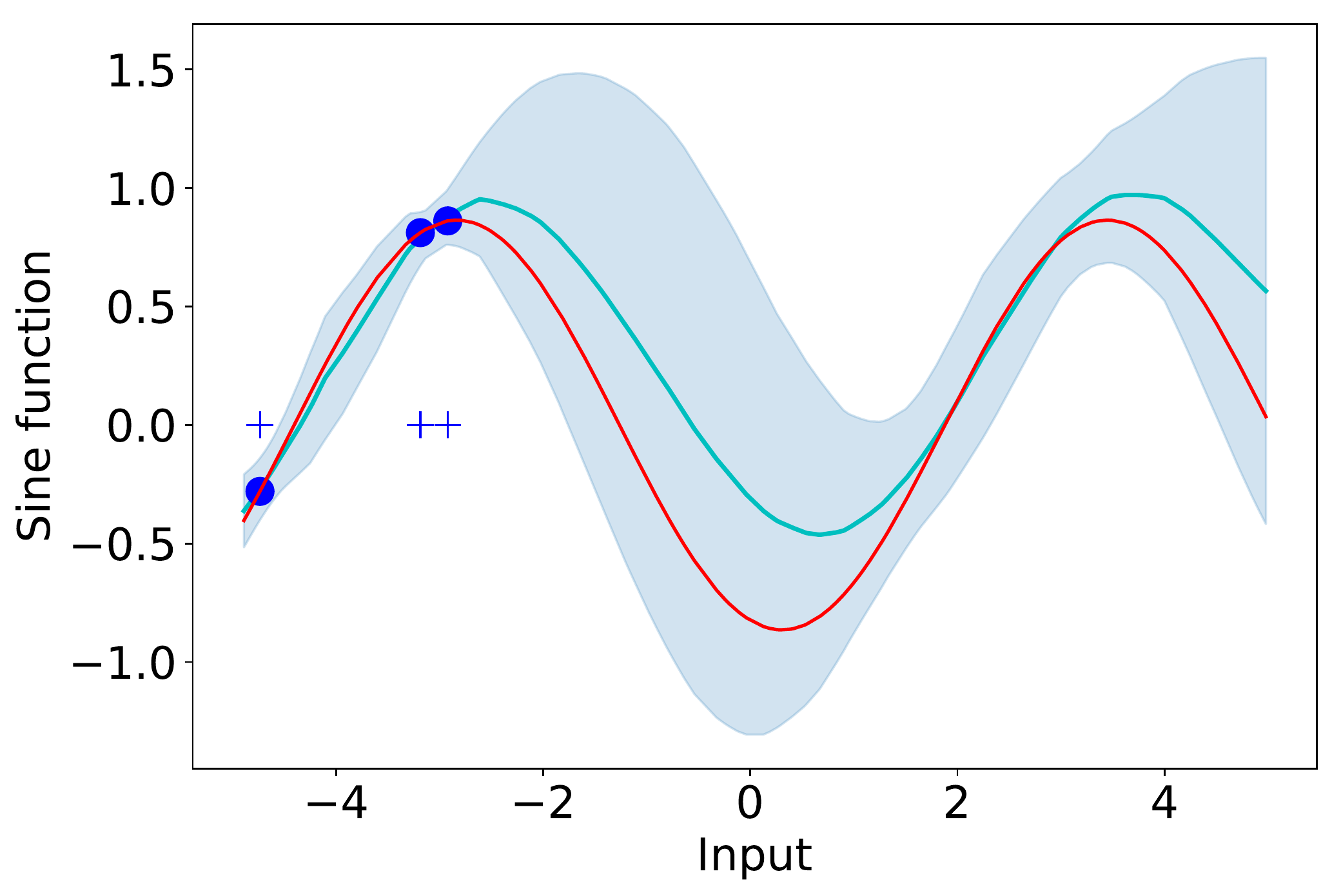}} \\
    \vspace*{-2pt}
    $K=4$ & $K=5$ & $K=7$ \\
    \vspace*{-2pt}
    {\includegraphics[scale=0.22]
    {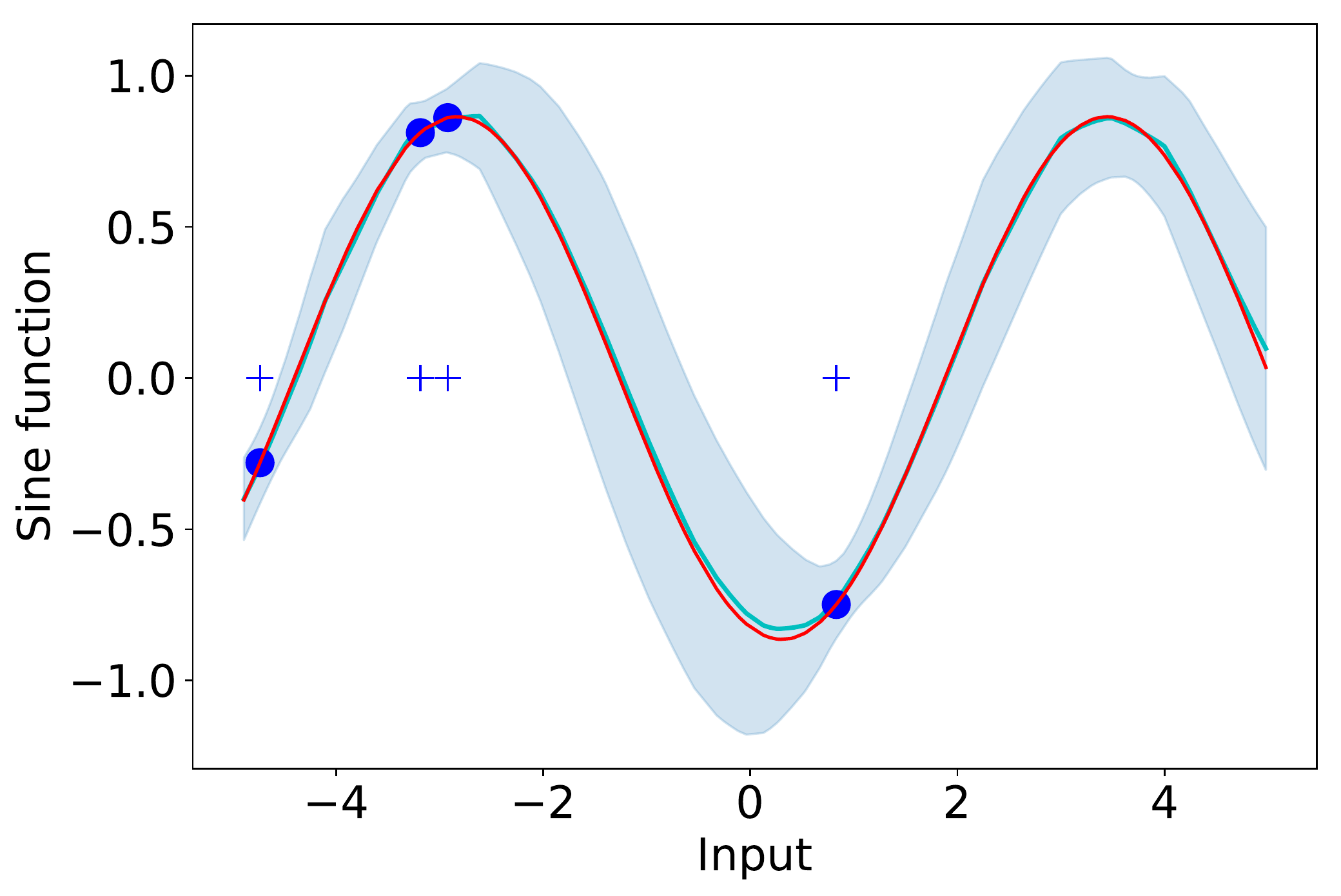}} &
    {\includegraphics[scale=0.22]
    {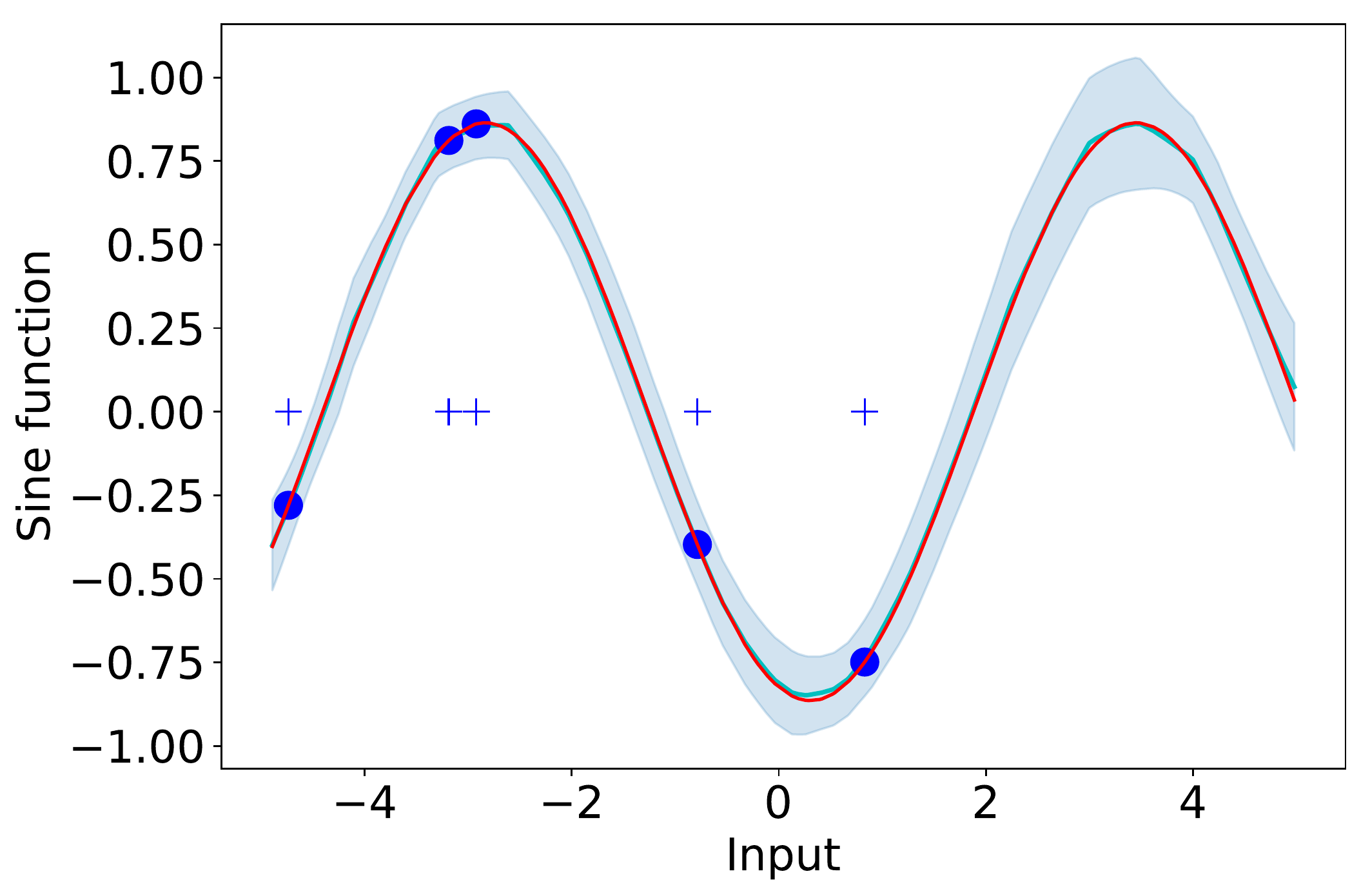}} &
    {\includegraphics[scale=0.22]
    {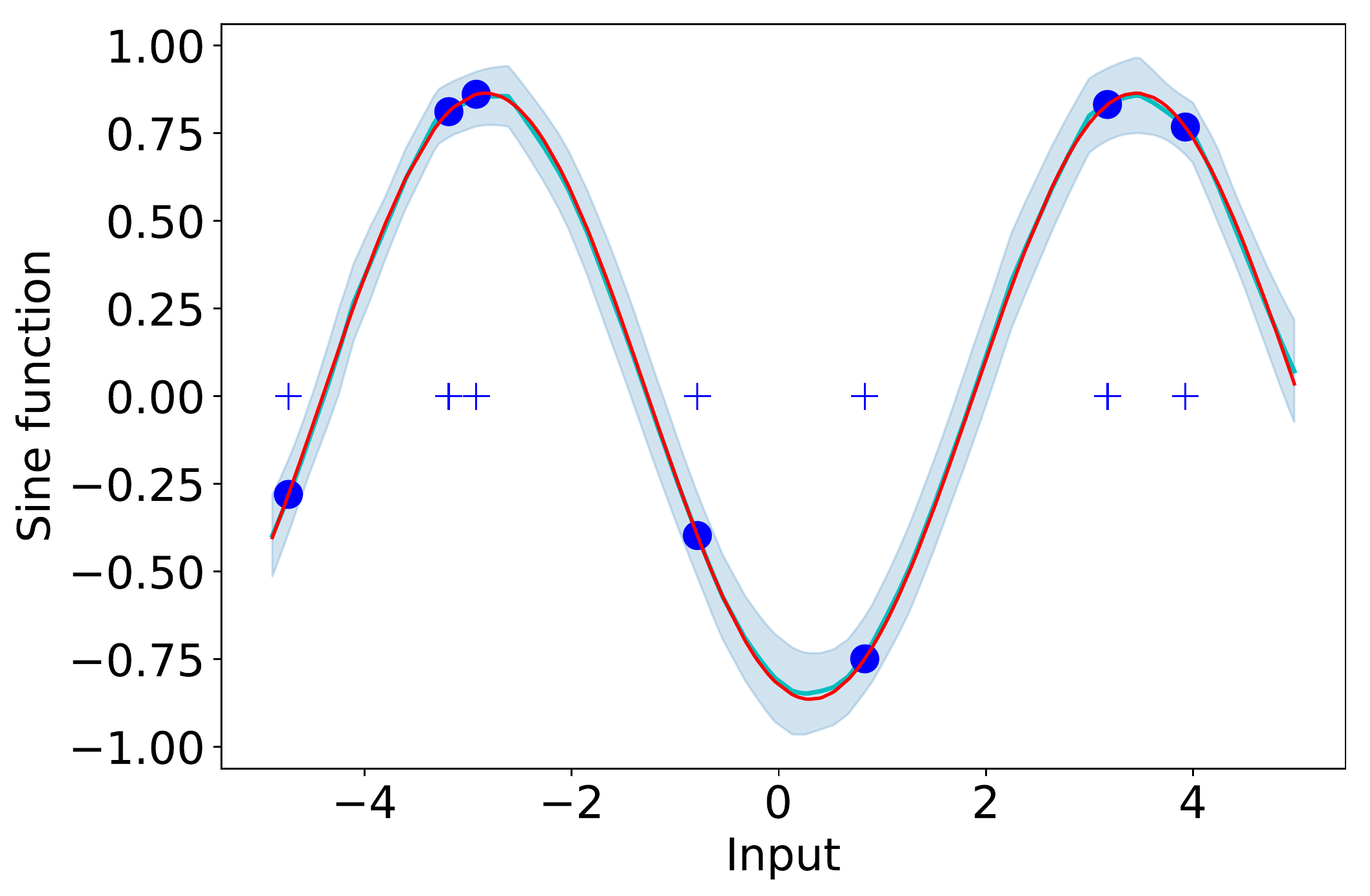}} 
    \end{tabular}
    \vspace*{-9pt}
    \caption{%
        Comparison of the ground truth sinusoid function (red) with the mean prediction of GP-VIB (light blue). As the number of shots $K$ increases (the observations are depicted by blue points), the uncertainty (shaded area) of the GP decreases and its mean predictions become closer to the ground truth. With just $K=4$ points, the mean prediction of the GP closely matches the sinusoid function.
    } 
    \label{fig:sines}
\end{figure*}

{\bf Sinusoid regression.} We first evaluate the method on sinusoid regression, following the settings of \citet{finn17a}. Each task involves regressing from the input to the output of a sine wave, where the amplitude and phase of the sinusoids vary across tasks.

We fit GP-VIB with regularization $\beta=1$ and linear kernel. More in detail, the kernel feature vector $\feat(x;\theta)$ is obtained by the last hidden layer of the same architecture used in MAML \citep{finn17a}. Based on this $M$-dimensional feature vector $\feat(x;\theta)$, we obtain the kernel as $k_\theta(x,x') = \frac{1}{M} \feat(x;\theta)^\top \feat(x';\theta)$ (where the kernel variance $\sigma_f^2$ is fixed to $1/M$).
Besides GP-VIB, we also fit MAML (with varying number of adaptation steps) for comparisons.
In particular, following \cite{finn17a}, we meta train MAML with one gradient adaptation step, while at meta testing we consider several adaptation steps $(1,5,10)$.

We report the performance of both methods in terms of $K$-shot mean squared error (MSE) in \Cref{table:few_shot_regression_one}. We observe that GP-VIB significantly outperforms MAML, especially as $K$ grows. \Cref{fig:sines} explains this few-shot predictive ability of GP-VIB, as the GP posterior uncertainty decreases as $K$ grows, and the posterior mean becomes very close to the ground truth sinusoid after only $K=4$ shots.

\begin{table}[t]
    \setlength\tabcolsep{3pt}
    {
    \caption{%
    MSE for $K$-shot sinusoid regression (mean values and $95\%$ confidence intervals after $10$ repeats). GP-VIB exhibits significantly smaller error than MAML.
    }
    \label{table:few_shot_regression_one}}
    {\small
    \begin{tabular}{@{}lccc@{}}
    \toprule
    \textbf{Method} & $K=5$ & $K=10$ & $K=20$ \\
    \midrule
        MAML-steps 1 & $0.600 {\pm 0.662}$ & $0.359 {\pm 0.015}$ & $0.228 {\pm 0.018}$ 
        \\ 
        MAML-steps 5 & $0.311  {\pm 0.013}$ & 
        $0.12  {\pm 0.006}$ & 
        $0.06  {\pm 0.004}$ 
        \\
        MAML-steps 10 & $0.280 {\pm 0.013}$ &
        $0.096 {\pm 0.005}$ &
        $0.043 {\pm 0.003}$
     \\   
    GP-VIB [ours] & $\mathbf{0.02 {\pm 0.014}}$ & 
    $\mathbf{0.002 {\pm 0.001}}$ & $\mathbf{0.001  {\pm 0.001}}$ \\
    \bottomrule
    \end{tabular}
    }
    \vspace*{-5pt}
\end{table}

{\bf Few-shot classification.}
We now evaluate GP-VIB on few-shot classification. To provide a fair comparison across different methods, we follow the unified training and evaluation protocol of \citet{patacchiola2019deep} and implement GP-VIB building on their PyTorch code.\footnote{%
    The code by \citet{patacchiola2019deep} builds on the implementation of \citet{chen2018a} and is available at \url{https://github.com/BayesWatch/deep-kernel-transfer}.
}

We train both $1$-shot and $5$-shot versions of GP-VIB in four settings:
Caltech-UCSD Birds (CUB) \citep{WahCUB_200_2011}, mini-ImageNet, and two cross-domain transfer tasks---training on mini-ImageNet and testing on CUB, and training on Omniglot \citep{Lake2011OneSL} and testing on EMNIST \citep{cohen2017emnist}.
The CUB dataset consists of $11788$ images across $200$ classes. We divide the dataset in $100$ classes for training, $50$ for validation, and $50$ for testing \citep{chen2018a}. The mini-ImageNet dataset  consists of a subset of $100$ classes ($600$ images for each class) taken from the ImageNet dataset. We use $64$ classes for training, $16$ for validation
and $20$ for testing, as is common practice \citep{Ravi2017OptimizationAA,chen2018a}. The Omniglot dataset contains $1623$ characters taken from $50$ different languages. Following the experimental protocol of \citet{nichol2018firstorder}, the number of classes is increased to $6492$ by applying data augmentation and adding examples rotated by 90 degrees. We use $4114$ classes for training. The EMNIST dataset  contains single digits and characters from the
English alphabet. The total $62$ classes are divided into $31$ for validation and $31$ for test. The whole experimental protocol is taken from \cite{patacchiola2019deep}.

\begin{table*}[th]
    \centering
    \caption{Average accuracy and standard deviation (percentage, based on $3$ independent runs) on few-shot classification ($5$-ways). The two methods with the best average accuracy are highlighted in bold.
    All results, except for GP-VIB, are taken from \citet{patacchiola2019deep}. GP-VIB provides competitive accuracy, obtaining state-of-the-art results in some cases.
    \vspace*{-3pt}
    }
    \label{tab:classif_accuracy}
    \begin{subtable}{\textwidth}
        \centering
        {\small
        \begin{tabular}{lcccc}
        \toprule
        &  \multicolumn{2}{c}{\bfseries  CUB} & 
        \multicolumn{2}{c}{\bfseries  mini-ImageNet} \\
        \bfseries  Method & \bfseries  1-shot & \bfseries
        5-shot & \bfseries  1-shot & \bfseries 5-shot \\
          \midrule %
        Feature Transfer & $46.19 \pm 0.64$ & $68.40  \pm 0.79$ & $39.51  \pm 0.23$ & $60.51  \pm 0.55$\\
        Baseline++ \citep{chen2018a} & $61.75  \pm 0.95$ & $\mathbf{78.51  \pm 0.59}$ & $47.15  \pm 0.49$ & $\mathbf{66.18  \pm 0.18}$ \\
        MatchingNet \citep{Vinyals16} & $60.19  \pm 1.02$ & $75.11  \pm 0.35$ & $48.25  \pm 0.65$ & $62.71  
        \pm 0.44$\\
        ProtoNet \citep{snell2017} & $52.52  \pm 1.90$ & $75.93  \pm 0.46$ & $44.19  \pm 1.30$ & $64.07  \pm 0.65$ \\
        MAML \citep{finn17a} & $56.11  \pm 0.69$ & $74.84  \pm 0.62$ & $45.39  \pm 0.49$ & $61.58  \pm 0.53$ \\
        RelationNet \citep{sung2018learning} & $62.52  \pm 0.34$ & $78.22  \pm 0.07$ & $48.76  \pm 0.17$ &  $64.20  \pm 0.28$ \\
        DKT + Linear \citep{patacchiola2019deep}  & $60.23\pm 0.76$ & $74.74 \pm 0.22$ & $48.44 \pm 0.36$ & $62.88 \pm 0.46$ \\
        DKT + CosSim \citep{patacchiola2019deep}  & $\mathbf{63.37  \pm 0.19}$ & $77.73  \pm 0.26$ & $48.64  \pm 0.45$ & $62.85  \pm 0.37$ \\
        DKT + BNCosSim \citep{patacchiola2019deep} &  $\mathbf{62.96  \pm 0.62}$ & $77.76  \pm 0.62$ & $\mathbf{49.73  \pm 0.07}$ & $64.00  \pm 0.09$ \\ \midrule
        GP-VIB + Linear [ours] & $62.73 \pm 0.59$ & $77.78 \pm 0.27$ & $49.61 \pm 0.29$ & $\mathbf{65.84 \pm 0.22}$  \\
        GP-VIB + CosSim [ours] & $62.36 \pm 0.77$ & $\mathbf{78.35 \pm 0.23}$ & $\mathbf{49.80 \pm 0.52}$ & $65.60 \pm 0.54$ \\
        \bottomrule %
        \end{tabular}
        }
        \caption{Few-shot classification on CUB and mini-ImageNet.}
        \label{tab:classif_accuracy_cub_miniImagenet}
    \end{subtable}
    \begin{subtable}{\textwidth}
        \centering
        {\small
        \begin{tabular}{lcccc}
        \toprule
        &  \multicolumn{2}{c}{\bfseries Omniglot $\rightarrow $ EMNIST}  &   \multicolumn{2}{c}{\bfseries mini-ImageNet $\rightarrow$ CUB} \\
        \bfseries  Method & \bfseries  1-shot & \bfseries  
        5-shot & \bfseries  1-shot & \bfseries 5-shot \\
          \midrule %
        Feature Transfer & $64.22 \pm 1.24$ & $86.10 \pm 0.84$ & $32.77 \pm 0.35$ & $50.34 \pm 0.27$ \\
        Baseline++ \citep{chen2018a} & $56.84 \pm 0.91$ & $80.01 \pm 0.92$ & $39.19 \pm 0.12$ & $\mathbf{57.31 \pm 0.11}$ \\
        MatchingNet \citep{Vinyals16} & $75.01 \pm 2.09$ & $87.41 \pm 1.79$ & $36.98 \pm 0.06$ & $50.72 \pm 0.36$ \\
        ProtoNet \citep{snell2017} & $72.04 \pm 0.82$ & $87.22 \pm 1.01$ & $33.27 \pm 1.09$ & $52.16 \pm 0.17$ \\
        MAML \citep{finn17a} & $72.68 \pm 1.85$ & $83.54 \pm 1.79$ & $34.01 \pm 1.25$ & $48.83 \pm 0.62$ \\
        RelationNet \citep{sung2018learning} & $75.62 \pm 1.00$ & $87.84 \pm 0.27$ & $37.13 \pm 0.20$ & $51.76 \pm 1.48$\\
        DKT + Linear \citep{patacchiola2019deep} & $\mathbf{75.97 \pm 0.70}$ & $89.51 \pm 0.44$ & $38.72 \pm 0.42$  & $54.20 \pm 0.37$ \\
        DKT + CosSim \citep{patacchiola2019deep} & $73.06 \pm 2.36$ & $88.10 \pm 0.78$ & $\mathbf{40.22 \pm 0.54}$ & $55.65 \pm 0.05$\\
        DKT + BNCosSim \citep{patacchiola2019deep} & $75.40 \pm 1.10$ & $\mathbf{90.30 \pm 0.49}$ & $40.14 \pm 0.18$ & $56.40 \pm 1.34$\\ \midrule
        GP-VIB + Linear [ours] & $\mathbf{76.01 \pm 0.54}$ & $\mathbf{89.93 \pm 0.23}$ & $39.52 \pm 0.7$ & $54.51 \pm 0.45$ \\
        GP-VIB + CosSim [ours] & $74.58 \pm 1.56$ & $88.73 \pm 0.75$ & $\mathbf{40.70 \pm 0.48}$ & $\mathbf{56.70 \pm 0.62}$ \\
        \bottomrule %
        \end{tabular}
        }
        \caption{Cross-domain few-shot classification.}
        \label{tab:classif_accuracy_cross_domain}
    \end{subtable}
    \vspace*{-6pt}
\end{table*}

To fit GP-VIB, we employ the commonly used Conv$4$ architecture with $64$ channels \citep{Vinyals16} and we use the Adam optimizer with the default learning rate of $10^{-3}$.
We use the simplified GP encoder described in \Cref{sec:amortisationq} and we set the regularization coefficient $\beta=0.001$ in all our experiments. We may obtain better results via hyperparameter search for each specific dataset, but using a single value for all of them showcases the robustness of the method. The kernel feature vector $\feat(x;\theta)$ is obtained by the last hidden layer of the same architecture used in MAML. We consider two kernel functions: the standard linear kernel (labeled ``GP-VIB + Linear''), $k_\theta(x,x') = \frac{\exp(v)}{M} \feat(x;\theta)^\top \feat(x';\theta)$ and the cosine similarity kernel (``GP-VIB + CosSim''), $k_\theta(x,x') =
\frac{ \exp(v) \feat(x;\theta)^\top \feat(x';\theta)}{|| \feat(x;\theta)||\; ||\feat(x';\theta)||}$, where the out-scale parameter $v$ is updated with a smaller learning rate ($10^{-4}$), exactly as in \cite{patacchiola2019deep}.

We compare GP-VIB against a series of baseline methods, including MAML as well as more recent methods. We evaluate each method in terms of the classification accuracy. We report the results in \Cref{tab:classif_accuracy}, together with the standard deviation obtained from three independent runs. We can observe that GP-VIB is a competitive method, and it exhibits state-of-the-art results in some cases. In the settings where GP-VIB is not the best performing method, its accuracy is close to the best one.
\section{Conclusions}

We have introduced an information theoretic framework for meta learning by using a variational approximation \citep{Alemietal2017,Chalk16,aless2016information} to the information bottleneck \citep{tishby99information}.
Based on this VIB view, we have developed a memory-based meta learning method that uses GPs to obtain a non-parametric stochastic encoding representation. We have shown experimentally that this method outperforms MAML in few-shot sinusoid regression and provides competitive performance on four few-shot classification problems, where it gives state-of-the-art results in some cases.

While we have demonstrated our method in few-shot
regression and classification, we believe that the scope of the information bottleneck for meta learning is much broader. For instance, a promising topic for future research is to consider applications in reinforcement learning.

\begin{acknowledgements}
    We are grateful to Yutian Chen for his help with the implementation and for many insightful discussions and comments on the manuscript. We also thank Jakub Sygnowski for technical advice regarding the implementation, and Yee Whye Teh and Matthias Bauer for their useful comments.
\end{acknowledgements}

\bibliography{main}

\clearpage

\appendix

\section{Further details about VIB in meta learning }
\label{app:VIB}

\subsection{Bounds on the mutual information}
\label{app:boundsonMI}

Here, we review the standard variational bounds on the mutual information from \citet{barberagakov03}. Recall the definition of the mutual information,
\begin{align*}
I(x,y) & = \int q(x,y) \log 
\frac{q(x,y)}
{q(x)q(y)} d x dy
\\& =
\int q(x,y) \log 
\frac{q(x|y)}
{q(x)} d x dy.
\end{align*}
By introducing $p(x|y)$ that approximates 
$q(x|y)$ we get
\begin{align*}
&I(x,y) =
\int q(x,y) \log 
\frac{p(x|y) q(x|y)}
{p(x|y) q(x)} d x dy\\
&= \!\int \!\!q(x,y) \log 
\frac{p(x|y)}
{q(x)} d x dy \!+\!\!
\int\!\! q(y) \text{KL}
[q(x|y) || p(x|y)] d y,
\end{align*}
which shows that
\begin{equation}
I(x,y) \geq  \int q(x,y) \log 
\frac{p(x|y)}
{q(x)} d x d y,
\label{eq:lowerboundmutual}
\end{equation}
since 
$\int q(y) \text{KL} [q(x|y) || p(x|y)] d y$ is non 
negative.

An upper bound is obtained similarly. Suppose 
$p(x)$ approximates $q(x)$; then 
\begin{align*}
I(x,y) & =
\int q(x,y) \log 
\frac{p(x) q(x|y)}
{p(x) q(x)} d x dy \\
&= \int q(x,y) \log 
\frac{q(x|y)}
{p(x)} d x dy -
\text{KL}
[q(x) || p(x)] d y,
\end{align*}
which shows that
\begin{equation}
I(x,y) \leq  \int q(x,y) \log 
\frac{q(x|y)}
{p(x)} d x dy.
\label{eq:upperboundmutual}
\end{equation}

\subsection{The general VIB meta learning case} 
\label{sec:generalcase}

Consider the general case, where we work with the unconditional mutual information and we wish to approximate the
information bottleneck (IB): $I(Z,\data^\val) - \beta I(Z, \data^\tr)$. Recall that the joint distribution is written as 
\begin{equation}
q_w(\data^\val, \data^\tr, Z ) =
q_w(Z| \data^\tr) p(\data^\val, \data^\tr),
\label{eq:joint2}
\end{equation}
from which we can express any marginal or conditional. In particular observe that  
$$
q_w(Z, D^\val) =
\int q_w(Z| \data^\tr) p(\data^\val, \data^\tr) 
d \data^\tr.
$$
If we have a function $f(Z,D^\val)$ and 
we wish to approximate the expectation, 
\begin{align}
& \int q_w(Z, \data^\val) f(Z,\data^\val) d Z d 
\data^\val \label{eq:expDvZ} \\
& = \int q_w(Z| \data^\tr) p(\data^\val, \data^\tr) f(Z,D^\val) d Z d \data^\val d \data^\tr, \nonumber
\end{align}
then, given that we sample a task pair 
$(\data^\val_i, \data^\tr_i) \sim
p(\data^\val, \data^\tr)$,
we can obtain the following unbiased estimate of this 
expectation,
\begin{equation}
\int q_w(Z| \data^\tr_i)  f(Z,\data^\val_i) d Z.
\label{eq:unbiasexpDvZ}
\end{equation}
We are going to make use of \Cref{eq:expDvZ,eq:unbiasexpDvZ} in the derivation below. 

To compute the variational approximation to IB, we need to lower bound
$I(Z, \data^\val)$ as
\begin{align}
& I (Z, \data^\val) 
= \int q_w(Z, \data^\val) 
\log 
\frac{q_w(Z, \data^\val)}
{ q_w(Z) p(\data^\val) } \nonumber \\
& = \int q_w(Z, \data^\val) 
\log 
\frac{q_w(\data^\val| Z)}
{p(\data^\val ) } d Z d \data^\val
\nonumber \\
& \geq \int q_w(Z, \data^\val) 
\log 
\frac{p_\theta(\data^\val| Z)}
{p(\data^\val )} d Z d \data^\val  
\quad  \text{(by  \Cref{eq:lowerboundmutual})}
\nonumber  \\
&  = \int q_w(Z, \data^\val) 
\log  p_\theta(\data^\val| Z) d Z d \data^\val  
+ \mathcal{H}(\data^\val), \nonumber 
\end{align}
where the entropy $\mathcal{H}(\data^\val)$  is just a constant. 

Subsequently, we upper bound  $I(Z, \data^\tr)$ 
as follows,
\begin{align}
& I(Z, \data^\tr) 
= \int q_w(Z, \data^\tr) 
\log 
\frac{q_w(Z, \data^\tr)}
{ q_w(Z) p(\data^\tr) } d Z d \data^\tr  \nonumber \\
& = \int q_w(Z | \data^\tr) p(\data^\tr)  
\log 
\frac{q_w(Z | \data^\tr)}
{q_w(Z) } d Z d \data^\tr, \nonumber \\
& \leq \int q_w(Z | \data^\tr) p(\data^\tr) 
\log 
\frac{q_w(Z | \data^\tr)}
{p_\theta(Z) } d Z d \data^\tr
\;\; \text{(by \Cref{eq:upperboundmutual})} \nonumber
\end{align}
Then we obtain the overall loss, $\mathcal{F}(\theta, w) \leq \mathcal{L}_{\textrm{IB}}(w)$: 
\begin{align}
\mathcal{F}(\theta,w)  
= & \int q_w(Z, \data^\val) 
 \log  p_\theta(\data^\val| Z) d Z d \data^\val  \nonumber \\
& -  \beta \int q_w(Z | \data^\tr) p(\data^\tr)  
\log \frac{q_w(Z | \data^\tr)}
{p_\theta(Z) } d Z d \data^\tr, \nonumber 
\end{align}
where we dropped the constant entropic term $\mathcal{H}(\data^\val)$.
Therefore, given a set of task pairs $\{\data^\tr_i, \data^\val_i \}_{i=1}^b$, where each 
$(\data^\tr_i, \data^\val_i) \sim \widetilde{p}(\data^\val, \data^\tr)$, 
the objective function 
for learning $(\theta, w)$ becomes the empirical average, 
$\frac{1}{b} \sum_{i=1}^b 
\mathcal{F}_i(\theta,w)$, where
\begin{align}
\mathcal{F}_i(w,\theta) 
= & \int q_w(Z_i | \data^\tr_i) \log  p_\theta(\data^\val_i| Z_i) d Z_i \label{eq:vibobjective2} \\   
& -  \beta \int q_w(Z_i | \data^\tr_i)  
\log \frac{q_w(Z_i | \data^\tr_i)}
{p_\theta(Z_i) } d Z_i, \nonumber
\end{align}
where for the first term we made use of \Cref{eq:expDvZ,eq:unbiasexpDvZ} with 
$f(\data^\val,Z) =  \log  p_\theta(\data^\val | Z)$.

\subsection{The supervised meta learning VIB case}
\label{app:vibsuper}

For the supervised meta learning case the joint density can be written as 
\begin{align}
& q_w(\data^\val, \data^\tr, Z ) \nonumber \\
& =
q_w(Z| Y^\tr, X^\tr, X^\val) p(Y^\tr, Y^\val | X^\tr, X^\val)
p(X^\val, X^\tr) , \nonumber \\ 
& = q_w(Z| Y^\tr, X)
p(Y^\tr, Y^\val | X)
p(X),
\label{eq:jointsuper}
\end{align}
where $X = (X^\tr, X^\val)$ and 
the encoding distribution $q_w(Z| Y^\tr, X)$ 
could depend on all inputs $X$ but only on the training
outputs $Y^\tr$.
The derivation of the VIB objective is similar as the general case,
with the difference that now we approximate
the conditional 
information bottleneck 
$I(Z,Y^\val|X) - \beta I(Z,Y^\tr|X)$,
where we condition on the inputs $X$. 
In other words, both $I(Z,Y^\val|X)$ and $I(Z,Y^\tr|X)$ 
are conditional mutual informations, i.e., they have the form 
\begin{align*}
I(z,y| x) & = \!
\int \!q(x) \!\left[\int \! q(z, y | x)  
\log 
\frac{q(z, y| x)}
{ q(z|x) q( y | x)
} d z d y \right]\! d x \\
& = \int q(z, y, x)  
\log 
\frac{q(z, y| x)}
{ q(z|x) q( y | x)
} d z d y d x.
\end{align*}
We can lower bound $I(Z,Y^\val|X)$ 
as follows,  
\begin{align*}
& \int \!\! p(X) \!\!\left[ \!\int \! q_w(Z, Y^\val | X)  
\log 
\frac{q_w(Z, Y^\val| X)}
{ q_w(Z|X) p( Y^\val | X)
} d Z d Y^\val \! \right]\!\! d X
\nonumber \\
& = 
\int p(X) \int  
q_w(Z, Y^\val | X) 
\log 
\frac{q_w(Y^\val| Z, X)}
{p(Y^\val | X)} d Z d Y^\val d X
\nonumber \\
& \geq 
\int p(X) \int  
q_w(Z, Y^\val | X) 
\log 
\frac{p_\theta(Y^\val| Z, X)}
{p(Y^\val | X)} d Z Y^\val d X
\nonumber \\ 
& =
\int 
q_w(Z, Y^\val , X) 
\log 
\frac{p_\theta(Y^\val| Z, X)}
{p(Y^\val | X)} d Z d Y^\val d X \nonumber \\
& = 
\int 
q_w(Z, Y^\val, X) 
\log p_\theta(Y^\val| Z, X) d Z d Y^\val d X \\
& \quad -
\int p(Y^\val,X) 
\log p(Y^\val | X) d Y^\val d X.
\end{align*}
In the second line above, $q_w(Z|X)$ cancels, and in the third line we have applied \Cref{eq:lowerboundmutual}.
Note that 
$
- \int p(Y^\val,X) 
\log p(Y^\val | X) d Y^\val d X
$
is just a constant that does not depend on tunable parameters. Also  
\begin{equation}
q_w(Z, Y^\val, X) 
\!=\!\! \int\! q_w(Z| Y^\tr, X)
p(Y^\tr, Y^\val | X) 
p(X) d Y^\tr,
\end{equation}
so that if we have a task sample 
$ (Y^\tr_i, Y^\val_i, X_i) \sim
p(Y^\tr, Y^\val | X) 
p(X)$, an unbiased estimate of the 
expectation $\int 
q_w(Z, Y^\val, X) 
\log p_\theta(Y^\val| Z, X) d Z d Y^\val d X$
is given by
\begin{equation}
\int 
q_w(Z| Y^\tr_i, X_i) 
\log p_\theta(Y^\val_i| Z, X_i) d Z.  
\label{eq:unbiasedyX}
\end{equation}

We upper bound 
$I(Z,Y^\tr|X)$ as follows, 
\begin{align*}
& \int \!p(X)\! \left[ \!\int \!\!q_w(Z, Y^\tr | X)
\log 
\frac{q_w(Z, Y^\tr| X)}
{ q_w(Z|X) p( Y^\tr | X)
} d Z d Y^\tr \right] \! d X \nonumber \\ 
& = 
\int p(X) \left[ \int q_w(Z, Y^\tr | X)
\log 
\frac{q_w(Z | Y^\tr, X)}
{ q_w(Z|X)} d Z d Y^\tr \right] \! d X,  
\nonumber \\
& \leq
\int p(X) \int q_w(Z, Y^\tr | X)
\log 
\frac{q_w(Z | Y^\tr, X)}
{ p_\theta(Z|X)}  d Z d Y^\tr d X,
\nonumber \\
& = \int q_w(Z | Y^\tr, X) 
p(Y^\tr, X)
\log 
\frac{q_w(Z | Y^\tr, X)}
{p_\theta(Z|X)}  d Z d Y^\tr d X,
\nonumber
\end{align*}
Then we obtain the overall objective, 
\begin{align}
& \mathcal{F}(\theta,w)  
= \int 
q_w(Z, Y^\val, X) 
\log p_\theta(Y^\val| Z, X) d Z d Y^\val d X
\nonumber \\
& - \!\beta\! 
\int \!\! q_w(Z | Y^\tr, X) 
p(Y^\tr, X)
\log 
\frac{q_w(Z | Y^\tr, X)}
{ p_\theta(Z|X)}  d Z d Y^\tr d X,
\nonumber 
\end{align}
where $p( Y^\tr | X)$ cancels in the second line, we have used \Cref{eq:upperboundmutual} in the third line, and we have dropped the constant term.
Therefore, given a set of task pairs 
the objective becomes the empirical average, 
$\frac{1}{b} \sum_{i=1}^b 
\mathcal{F}_i(\theta,w)$, where
\begin{align}
\mathcal{F}_i(\theta,w)  
= & \int 
q_w(Z| Y^\tr_i, X_i) 
\log p_\theta(Y^\val_i| Z, X_i) d Z \label{eq:app-VIBsuper} \\
& -  \beta 
\int q_w(Z | Y^\tr_i, X_i) 
\log 
\frac{q_w(Z | Y^\tr_i, X_i)}
{ p_\theta(Z|X_i)}  d Z,
\nonumber
\end{align}
where we made use of \Cref{eq:unbiasedyX}.

\subsection{Connection with variational inference}
\label{app:connectVariaitional}

As mentioned in the main paper, the VIB for meta learning (where we consider for simplicity the general case from \Cref{sec:generalcase})
is similar to applying approximate variational 
inference to a certain joint model over the validation set,
$$
p_\theta(\data^\val| Z) p_\theta(Z), 
$$
where $p_\theta(\data^\val| Z)$ is the decoder model, $p_{\theta}(Z)$ 
a prior model over the latent variables  and where  
the corresponding marginal likelihood is 
$$ 
p(\data^\val) = \int p_\theta(\data^\val| Z) p_\theta(Z) d Z.
$$
We can lower bound the log marginal likelihood 
with a variational distribution $q_w(Z | \data^\tr)$ that depends 
on the training set $\data^\tr$,
\begin{align}
\mathcal{F}_{\beta=1}(w,\theta) = & \int 
q_w(Z | \data^\tr) \log  p_\theta(\data^\val | Z) d Z   \\
&  -  \int q_w(Z | \data^\tr)  
 \log \frac{q_w(Z | \data^\tr)}
 {p_\theta(Z) } d Z, \nonumber
\end{align}
which corresponds to the VIB objective with $\beta=1$. %

\section{Transductive and non-transductive meta learning} 
\label{app:transductive}

Here, we discuss how the transductive and non-transductive settings that appear in few-shot 
image classification \citep{bronskill2020tasknorm, finn17a, nichol2018firstorder}, due to the use of batch-normalization, can be interpreted  under our VIB framework by defining suitable encodings. 
We shall use MAML as an example, but the discussion 
is more generally relevant. 

The transductive case occurs when the concatenated support and validation/test inputs $X=(X^\tr,X^\val)$ of a single task (we ignore the task index $i$ to keep the notation uncluttered)
are used to compute batch-norm statistics (possibly at different stages) shared by all validation/test points, when predicting those points. 
For MAML this implies a deterministic parametric encoding, i.e., common to all individual validation inputs $x^\val_j \in X^\val$, obtained by a sequence 
of two steps: (i) Obtain first 
the task-specific parameter $\taskparams$ in the usual way by the support loss, i.e.,
$\taskparams = \theta + \Delta(\theta,\data^\tr)$. If batch-normalization is used here, then the statistics are computed only by $X^\tr$.  (ii)
Compute the validation loss by applying batch-normalization on $X^\val$ or the union 
$X= X^\tr \cup X^\val$ (the union seems to be a better choice, but not used often in practice for computational reasons; e.g., \cite{finn17a, nichol2018firstorder} prefer to use only $X^\val$). 
In both cases, the underlying encoder is parametric 
over the final effective  task parameter $\widetilde{\taskparams}= BN(\taskparams,X)$, where
$BN$ denotes the final batch-norm operation that outputs a parameter vector,
that predicts all validation points and it is a deterministic delta measure. 

In contrast, the non-transductive setting occurs when each individual validation input $x^\val_j$ 
is concatenated with the support inputs $X^\tr$ to 
form the sets $x^\val_j \cup X^\tr$, $j=1,\ldots,n^\val$. 
Then, each set $x^\val_j \cup X^\tr$ is used to compute
point-specific batch-norm statistics when predicting the corresponding validation output $y^\val_j$. 
Under the VIB framework this corresponds to a non-parametric encoding, which grows with the  size of the validation set. The first deterministic 
step of this encoder is the same (i) above from the transductive case but the second step differs in the sense that now we get a validation point-specific task parameter $\widetilde{\taskparams}_j = BN(\taskparams, x^\val_j \cup X^\tr)$
by computing the statistics using the set 
$x^\val_j \cup X^\tr$. 
For MAML, this encoding becomes, $Z \equiv \{\widetilde{\taskparams}_j \}_{j=1}^{n^\val}$, and the encoder distribution is a product of delta measures. i.e., $p(\{\widetilde{\taskparams}_j \}_{j=1}^{n^\val} | Y^\tr, X) \equiv \prod_{j=1}^{n^\val} \delta_{\widetilde{\taskparams}_j, BN(\theta +
\Delta(\theta,\data^\tr), x^\val_j \cup X^\tr)}$.

Finally, note that under the VIB perspective 
it does not make much sense to meta train 
transductively and meta test non-transductively 
and vice versa, since this changes the encoding. That is, in meta testing we should do the same as in meta training.

\section{Further details about the Gaussian process method}
\label{app:detailsgp}

For simplicity next we ignore the task index $i$ to keep the notation uncluttered, and write for example $\bff^\tr_i$ as $\bff^\tr$.

\subsection{Derivation of the VIB bound}
\label{app:dervibbound}

The VIB objective for a single task from \Cref{eq:app-VIBsuper} in the main paper is computed as follows 
\begin{align}
&  
\sum_{j=1}^{n^\val} \Exp_{q(f^\val_j)} [
\log  p(y^\val_j | f^\val_j)]
-  \beta \int p(\bff^\val |\bff^\tr, X^\val, X^\tr) q(\bff^\tr | \data^\tr) \nonumber \\
& \quad \times \log \frac{p(\bff^\val |\bff^\tr,  X^\val, X^\tr) q(\bff^\tr | \data^\tr) }
 {p(\bff^\val |\bff^\tr, X^\val, X^\tr) p(\bff^\tr | X^\tr)} d \bff^\tr d \bff^\val
   \nonumber \\ 
 &  
=\!\sum_{j=1}^{n^\val} \Exp_{q(f^\val_j)} [
\log  p(y^\val_j | f^\val_j)] 
 -  \!\beta\! \int \!\! q(\bff^\tr | \data^\tr) 
 \log \frac{q(\bff^\tr | \data^\tr) }
 {p(\bff^\tr | X^\tr)} d \bff^\tr
  \nonumber \\ 
 & 
=\!\sum_{j=1}^{n^\val} \Exp_{q(f^\val_j)} 
[\log  p(y^\val_j | f^\val_j) ] \!-\!  \beta 
\text{KL} \left[ q(\bff^\tr | \data^\tr) || p(\bff^\tr | X^\tr) \right],
\label{eq:appvibgpobjective}
\end{align}
where $q(f^\val_j) 
= \int p(f^\val_j |\bff^\tr, x^\val_j, X^\tr) q(\bff^\tr | \data^\tr) d \bff^\tr$ is a marginal Gaussian over an individual 
validation function value $f^\val_j$, as also explained
in the main paper. Specifically, $q(f^\val_j)$ depends 
on the training set $(Y^\tr,X^\tr)$ and the single validation input $x^\val_j$, so intuitively 
from the training set and the corresponding function 
values $\bff^\tr$ we extrapolate (through 
the conditional GP $p(f^\val_j |\bff^\tr, x^\val_j, X^\tr)$)
to the input $x^\val_j$ in order to predict its 
function value $f^\val_j$. 

Given the specific amortization of  $q(\bff^\tr | \data^\tr)$:
\begin{align}
& q(\bff^\tr | \data^\tr) 
= \frac{\left( \prod_{j=1}^{n^\tr} \mathcal{N}(m^\tr_j | %
s^\tr_j) \right)  \mathcal{N}(\bff^\tr |0, {\bf K}^\tr) }
{\mathcal{N}(\bfm^\tr |0, {\bf K}^\tr + {\bf S}^\tr)} \label{eq:app-qfDt} \\
& = \mathcal{N}(\bff^\tr
| {\bf K}^\tr(  {\bf K}^\tr + {\bf S}^\tr )^{-1} \bfm^\tr, 
{\bf K}^\tr - {\bf K}^\tr({\bf K}^\tr + {\bf S}^\tr)^{-1}  {\bf K}^\tr ),
 \nonumber
\end{align}
the VIB objective, by using the middle part of 
\Cref{eq:app-qfDt}, can be written in the following form, 
\begin{align}
& \sum_{j=1}^{n^\val} \Exp_{q(f^\val_j)}[ 
\log  p(y^\val_j | f^\val_j)] \! -  \!\beta \sum_{j=1}^{n^\tr} \Exp_{q(f^\tr_j)} 
[\log \mathcal{N}(m^\tr_j | 
f^\tr_j,s^\tr_j)] \nonumber  \\
& + \beta \log \mathcal{N}(\bfm^\tr |{\bf 0}, {\bf K}^\tr + {\bf S}^\tr), 
\label{eq:app-vibobjective_2}
\end{align}
which is convenient from computational and programming point of view.
Specifically, to compute this 
we need to perform a single Cholesky 
decomposition of ${\bf K}^\tr + {\bf S}^\tr$ which scales as $O((n^\tr)^3)$, i.e., cubically w.r.t.\ the size of the 
support set $n^\tr$. This is fine for small support sets (which is the standard case in few-shot learning)
but it can become too expensive when $n^\tr$ becomes very large. However, given that the kernel has the linear form 
$ k_{\theta}(x, x') =  \feat(x;\theta)^\top \feat(x';\theta)$ (ignoring any kernel variance $\sigma_f^2$ for notational simplicity), where $\feat(x_i;\theta)$ is $M$-dimensional and given that $M \ll n^\tr$, we can also carry out the computations based on the Cholesky decomposition 
of a matrix of size $M \times M$. This 
is because ${\bf K}^\tr = \Feat^\tr {\Feat^\tr}^\top$, where $\Feat^\tr$ is an $n^\tr \times M$ matrix  
storing as rows the features vectors on the support inputs $X^\tr$, and therefore we can apply the standard matrix inversion and determinant lemmas for the matrix $\Feat^\tr {\Feat^\tr}^\top + {\bf S}^\tr$  when computing $\log \mathcal{N}(\bfm^\tr |{\bf 0}, {\bf K}^\tr + {\bf S}^\tr)$. Such $O(M^3)$ computations also gives us the quantities $q(f^\val_j)$ and $q(f^\tr_j)$, as explained next.  

\subsection{Data efficient GP meta testing prediction with constant memory}
\label{app:constantmemory}

Once we have trained the GP meta learning system 
we can consider meta testing where a new fresh 
task is provided having a support set $\data^\tr_* = (Y^\tr_*, X^\tr_*)$ 
based on which we predict at any arbitrary
validation/test input $x_*$. This requires 
to compute quantities (such as the mean value 
$\Exp[y_*]$) associated with the predictive density
\begin{align*}
q(y_*) & = \int p(y_* | f_*) 
p(f_* |\bff^\tr_*, x_*, X^\tr_*) q(\bff^\tr_* | 
\data^\tr_*) d f_* d \bff^\tr_* \\
& = \int p(y_* | f_*) 
q(f_*) d f_*,
\end{align*}
where $q(f_*)$ is an univariate Gaussian given by  
$$
q(f_*)
\!=\! \mathcal{N}(f_*| \bk_*^t 
({\bf K}^\tr + {\bf S}^\tr)^{-1} \bfm^\tr
,k_* 
-  \bk_*^t ({\bf K}^\tr + {\bf S}^\tr)^{-1} {\bk_*^t}^\top \!), 
$$
$$
\bk_*^t = \feat_*^\top
\Feat^\tr, \ {\bf K}^\tr = \Feat^\tr {\Feat^\tr}^\top, \ 
k_* = \feat_*^\top \feat_*, 
\ \feat_* = \feat(x_*;\theta).
$$
Here, $\Feat^\tr$ is an $n^\tr_* \times M$ matrix   
storing as rows the features vectors on the support inputs $X^\tr_*$. Note that if we wish to evaluate $q(y_*)$ at certain value of $y_*$, and given that the likelihood $p(y_* | f_*)$ is not 
the standard Gaussian, we can use 1-D 
Gaussian quadrature or Monte Carlo by sampling 
from $q(f_*)$.

An interesting property of the above predictive density is that when the support 
set $\data^\tr_*$ can grow 
incrementally, e.g., individual data points or mini-batches are added sequentially,  
the predictive density can be implemented 
with constant memory without requiring to explicit memorize the points in the support. The reason is that the feature parameters 
$\theta$ remain constant at meta test time and 
the kernel function is linear, so we can apply standard tricks to update the sufficient statistics as in Bayesian linear regression.  

More precisely, what we need to show is that 
we can sequentially update the mean and variance of $q(f_*)$ with constant memory. The distribution 
$q(f_*)$ can be written as
\begin{align}
q(f_*)
& = \mathcal{N}\Big(f_*\Big|  \feat_*^\top
{\Feat^\tr}^\top
(\Feat^\tr {\Feat^\tr}^\top + {\bf S}^\tr)^{-1} \bfm^\tr
, \\
& \qquad\qquad\feat_*^\top \left( I 
- 
{\Feat^\tr}^\top (\Feat^\tr {\Feat^\tr}^\top + {\bf S}^\tr)^{-1} \Feat^\tr
\right)
\feat_*\Big) 
\nonumber \\
& = \mathcal{N}\Big(f_*\Big| \feat_*^\top
({\Feat^\tr}^\top [{\bf S}^\tr]^{-1} \Feat^\tr + I)^{-1} {\Feat^\tr}^\top [{\bf S}^\tr]^{-1 }\bfm^\tr
,\nonumber\\
&\qquad\qquad \feat_*^\top ({\Feat^\tr}^\top [{\bf S}^\tr]^{-1} \Feat^\tr + I)^{-1} \feat_* \Big), \nonumber
\end{align}
where we applied the matrix inversion lemma backwards to write  
$I - {\Feat^\tr}^\top (\Feat^\tr {\Feat^\tr}^\top + {\bf S}^\tr)^{-1} \Feat^\tr = ({\Feat^\tr}^\top [{\bf S}^\tr]^{-1} \Feat^\tr + I)^{-1}$ and also used that 
${\Feat^\tr}^\top
(\Feat^\tr {\Feat^\tr}^\top + {\bf S}^\tr)^{-1} 
= {\Feat^\tr}^\top (\Feat^\tr {\Feat^\tr}^\top [{\bf S}^\tr]^{-1} + I)^{-1} [{\bf S}^\tr]^{-1} = ({\Feat^\tr}^\top [{\bf S}^\tr]^{-1} \Feat^\tr + I)^{-1} {\Feat^\tr}^\top [{\bf S}^\tr]^{-1}$ (based on the identity 
$(A B + I)^{-1} A = A (B A + I)^{-1}$). Now observe that the $M$-dimensional  vector $\bb^\tr
= {\Feat^\tr}^\top  [{\bf S}^\tr]^{-1} \bfm^\tr 
= \sum_{j=1}^{n^\tr} \feat(x^\tr_j;\theta) 
\frac{m^\tr_j}{s^\tr_j}$ can grow incrementally without memorizing the feature vectors $\feat(x^\tr_j;\theta)$ based on the recursion 
$\bb^\tr \leftarrow \bb^\tr 
+ \feat(x^\tr_j;\theta) 
\frac{m^\tr_j}{s^\tr_j}$ (with the initialization $\bb^\tr = 0$) as individual data points (similarly for mini-batches) 
are added in the support set: $\data^\tr \leftarrow \data^\tr \cup (x^\tr_j, y^\tr_j)$. Similarly, the $M \times M$ matrix 
$A^\tr = {\Feat^\tr}^\top  [{\bf S}^\tr]^{-1} \Feat^\tr
= \sum_{j=1}^{n^\tr} \frac{1}{s^\tr_j} \feat(x^\tr_j;\theta) \feat(x^\tr_j;\theta)^\top$
can also be computed recursively with  
constant $O(M^2)$ memory.

Finally, note that the above constant memory 
during meta testing can only be implemented when the feature vector $\theta$ is fixed.

\subsection{Multi-class classification}
\label{app:multiclass}

For multi-class classification meta 
learning problems we need to introduce  
as many latent functions as classes. 
For instance, when the number of classes for each task is $N$ we will need 
$N$ latent functions $f_n(x)$ which all are  independent draws from the same GP. 
The marginal GP prior on the training and validation function values for a certain task factorizes as 
$$
\prod_{n=1}^N p(\bff^\val_n |\bff^\tr_n,  X^\val, X^\tr) p(\bff^\tr_n | X^\tr). 
$$
We assume a factorized encoding distribution of the form
$$
\prod_{n=1}^N p(\bff^\val_n |\bff^\tr_n,  X^\val, X^\tr) q(\bff^\tr_n | \data^\tr),
$$
where each 
\begin{align*}
& q(\bff^\tr_n | \data^\tr)\\
&= 
\mathcal{N}(\bff^\tr_n
| {\bf K}^\tr(  {\bf K}^\tr + {\bf S}^\tr )^{-1} \bfm^\tr_n, 
{\bf K}^\tr - {\bf K}^\tr({\bf K}^\tr + {\bf S}^\tr)^{-1}  {\bf K}^\tr).
\end{align*}
Here, 
$\bfm^\tr_n = Y^\tr_n \circ 
\widetilde{\bfm}^\tr$, and $Y^\tr_n$ is a vector obtaining the value $1$ for each data point $x^\tr_j$ that belongs to class $n$  and $-1$ otherwise. Note that the encoding distributions
share the covariance matrix and they only have 
different mean vectors.
The representation of  $\bfm^\tr_n$ 
makes the full encoding distribution permutation invariant to the values of the class labels. Since also  we are using shared (i.e., independent of class labels) amortized 
functions $\widetilde{m}_w(x)$ and $s_w(x)$, 
the terms $({\bf S}^\tr, \widetilde{\bfm}^\tr)$ 
are common to all $N$ factors. 
This allows to compute the VIB 
objective very efficiently (in way that is fully scalable w.r.t.\ the number of classes $N$)
by requiring only a single 
Cholesky decomposition of 
the matrix ${\bf K}^\tr + {\bf S}^\tr$. Specifically, 
by working similarly to \Cref{app:dervibbound}
we obtain the VIB objective per single task,  
\begin{align}
& \sum_{j=1}^{n^\val} \Exp_{q(\{f^\val_{n,j}\}_{n=1}^N)}[ 
\log  p(y^\val_j | \{f^\val_{n,j}\}_{n=1}^N )]  \nonumber \\
& -  \beta \sum_{n=1}^N \sum_{j=1}^{n^\tr} \Exp_{q(f^\tr_{n,j})} 
[\log \mathcal{N}(m^\tr_{n,j} | 
f^\tr_{n,j},s^\tr_j)] \nonumber \\  
& + 
\beta \sum_{n=1}^N \log \mathcal{N}(\bfm^\tr_n |{\bf 0}, {\bf K}^\tr + {\bf S}^\tr), 
\nonumber 
\label{eq:app-vibobjective_multiclass}
\end{align}
where $q(\{f^\val_{n,j}\}_{n=1}^N) = \prod_{n=1}^N q(f^\val_{n,j})$ and each univariate Gaussian $q(f^\val_{n,j})$ is given by the same expression 
as provided in \Cref{app:constantmemory}. The last two terms of the bound (i.e., the ones multiplied by the hyperparameter $\beta$) are clearly analytically computed,
while the first term involves an expectation 
of a log softmax since the likelihood is  
$$
p(y^\val_j = n | \{ f^\val_{n',j} \}_{n'=1}^N) 
= \frac{e^{f^\val_{n,j}}}
{\sum_{n'=1}^N e^{f^\val_{n',j}}}.
$$
To evaluate this expectation we apply first 
the reparametrization trick to move all tunable
parameters of $q(\{f^\val_{n,j}\}_{n=1}^N)$ inside the log-likelihood (so that we get a new expectation under a product of $N$ univariate standard normals) and then we apply Monte Carlo 
by drawing $200$ samples. 

Finally, note that to compute the predictive 
density we need to evaluate, 
$$
q(y_*) = \Exp_{q(\{ f_{n,*} \}_{n=1}^N) } \left[ p(y_* | \{f_{n,*}\}_{n=1}^N) \right], 
$$
which again is done by applying Monte Carlo 
by drawing $200$ samples from 
$q(\{ f_{n,*} \}_{n=1}^N)$.
To decide the classification label 
based on the maximum class predictive probability
(in order to compute, e.g., accuracy scores), we take advantage of the fact that all $N$ univariate predictive Gaussians $q(f_{n,*})$ have the same variance but different means, thus the predicted 
class can be equivalently obtained by taking the 
argmax of the means of these $N$ distributions.

\subsection{Specific GP implementation and amortization for few-shot classification}

For all few-shot multi-class classification experiments in order to implement the GP-VIB %
method we need to specify the feature vector $\feat(x;\theta)$ 
and the amortized variational functions 
$\widetilde{m}_w(x)$ and $s_w(x)$. The feature 
vector is specified to have exactly the same neural 
architecture used in previous works for all  datasets. %
Note that when 
computing the GP kernel function,
the feature vector $\feat(x;\theta)$ is also augmented with the value $1$ to automatically account for a bias term.

Regarding the two amortized variational functions needed to obtain the encoder, we   
consider a shared (with the GP functions) 
representation by adding two heads to the same feature vector $\feat(x;\theta)$: the first head corresponds to a linear output function $\widetilde{m}_w(x)$ 
and the second applies at the end the softplus activation $s_w(x) = \log(1 + \exp(a(x)))$ (since 
$s_w(x)$ represents variance)
where the pre-activation $a(x)$ is obtained 
by a linear function of the feature vector. 
For numerical stability we also apply a final 
clipping by bounding these functions 
so that $\widetilde{m}_w(x) \in [-20,20]$ 
and $s_w(x) \in [0.001, 20]$.  
The bounds $-20$ and $20$ are almost never realized 
during optimization, so they are not so crucial, 
in contrast 
the lower bound $0.001$ on $s_w(x)$ is rather crucial 
regarding numerical stability since it ensures 
that the minimum eigenvalue 
of the matrix ${\bf K}^\tr + {\bf S}^\tr$ (i.e., the matrix we need to decompose using Cholesky) is bounded below by $0.001$.

For the simplified encoder where $(\widetilde{m}_w(x^\tr_j), s_w(x_j^\tr)) : = (\widetilde{m}, \sigma^2)$ we simply 
learn two independent
scalar parameters $(\widetilde{m},\sigma^2)$, where $\sigma^2 = \log(1 + \exp(a))$ and $a$ is the actual parameter optimised. For 
$(\widetilde{m}, \sigma^2)$ we use 
the same bounds mentioned above.

\end{document}